\documentclass{IEEEtaes}
\usepackage{amsmath,amsfonts}
\usepackage{algorithmic}
\usepackage{algorithm}
\usepackage{array}
\usepackage[caption=false,font=normalsize,labelfont=sf,textfont=sf]{subfig}
\usepackage{textcomp}
\usepackage{stfloats}
\usepackage{url}
\usepackage{verbatim}
\usepackage{multirow}
\usepackage{graphicx}
\usepackage{cite}
\usepackage[caption=false,font=normalsize,labelfont=sf,textfont=sf]{subfig}
\usepackage{color}
\usepackage{amsthm, amssymb,bm}
\usepackage{mathtools}
\usepackage{array}       
\usepackage{booktabs}    

\jvol{XX}
\jnum{XX}
\jmonth{XXXXX}
\paper{1234567}
\pubyear{2022}
\doiinfo{TAES.2022.Doi Number}

\begin{document}
\title{PILOT: Privileged Imitation Learning for End-to-End Motion Planning of Autonomous UAVs under Partial Observability}

\author{Qingrui Zhang}
\author{Feng Xue}
\author{Xiang Zhou}
\author{Chenghao Yu}
\affil{School of Aeronautics and Astronautics, Sun Yat-sen University (Shenzhen Campus), Shenzhen, China} 
 
\markboth{AUTHOR ET AL.}{SHORT ARTICLE TITLE}
\maketitle

\begin{abstract}
Autonomous navigation in cluttered environments is hampered by partial observability and dynamic constraints. This paper presents PILOT, a constraint-aware privileged imitation learning framework for vision-based end-to-end UAV motion planning under partial observability. The framework distills planning strategies from a computationally intensive optimal control expert into a student policy regularized toward safety and dynamic requirements via a dual-objective loss function. To mitigate partial observability, a spatiotemporal perception fusion module using a Temporal Convolutional Network (TCN) is developed to integrate historical depth images and odometry. This module infers task-relevant latent context from historical observations, enhancing spatial awareness beyond the instantaneous FOV without maintaining persistent map memory. A trajectory parameterization layer mapping network outputs to a structured trajectory, while enabling explicit continuity, dynamic-consistency, and obstacle soft penalties during training, encouraging constraint satisfaction for unseen observations without formal guarantees. Simulations on quadrotor and fixed-wing aircraft demonstrate that PILOT achieves performance comparable to the privileged expert while reducing computational overhead by over 80\%. 
Successful indoor and outdoor zero-shot deployment confirms the practical feasibility and cross-domain generalization of the planner. \footnote{Simulation and experiment videos can be found at \href{https://qr-zhang.github.io/PILOT/}{https://qr-zhang.github.io/PILOT/}.}.
\end{abstract}


\section{\textsc{Introduction}}
\IEEEPARstart{U}{nmanned} Aerial Vehicles (UAVs) have gained significant attention across diverse industries—including package delivery \cite{Xu2024TAES, saunders_autonomous_2024}, target tracking \cite{liu2025TAES, liu2025RAL}, and environmental monitoring \cite{Julian2019JGCD, Hong2023TAES}—owing to their exceptional maneuverability and versatile capabilities. As mission profiles evolve, autonomous navigation of UAVs must adapt to diverse scenarios. The primary challenge lies in the real-time generation of feasible trajectories under the stringent constraints of limited onboard perception and computation resources. This challenge is amplified in unstructured, cluttered low-altitude environments, where environmental complexity demands higher levels of autonomy \cite{Adhikari2020JGCD}. Autonomous flight in such scenarios necessitates a seamless integration of robust obstacle detection and dynamic motion planning \cite{mohta_fast_2018, Tang2024TAES, lu_autonomous_2025}. Vision-based systems are favored for this task due to their lightweight feature and rich 3D perception capabilities \cite{Kaiser2010TAES, Park2015TAES, Zhai202TAES}.

Conventional vision-based pipelines typically rely on modular architectures—separating perception, localization, and planning \cite{mohta_fast_2018, Tang2024TAES}. While transparent, these methods suffer from high integration complexity and cumulative error propagation across modules \cite{mohta_fast_2018}. They also exhibit limited adaptability to unexpected environmental changes or unstructured scenarios. To enhance performance and generality, optimization-based methods are often employed to minimize cost functions subject to environmental and dynamic constraints \cite{Tang2024TAES}. However, optimization-based methods are generally computationally intensive, a bottleneck that becomes increasingly acute when navigating non-convex obstacles in unstructured environments. Such overhead often precludes their viability for real-time onboard implementation on agile UAVs \cite{Eren2017JGCD, Zhang2022JGCD}. Furthermore, optimization-based methods remain modular by design, thus inheriting the issue of compounded error propagation across sequential stages. Independent of these architectural limitations, partial observability remains a persistent challenge for local sensing, as traditional planners frequently lack the spatiotemporal awareness necessary for safe navigation. Consequently, two fundamental research questions arise:
\begin{enumerate}
    \item  \emph{Can we develop a unified planner to eliminate the cumulative errors and computational overhead of modular optimization pipelines while maintaining high-performance trajectory generation?}
    \item \emph{Can UAVs overcome partial observability in cluttered environments to ensure safe navigation using local, egocentric observations?} 
\end{enumerate}

To address these challenges, learning-based planning, specifically Imitation Learning (IL), offers a promising path by shifting the heavy computational burden of optimization to an offline training phase \cite{DYZhang_RAL_2024, Song2023SR}, where learned policies serve as high-performance surrogates for real-time execution. Unlike Reinforcement Learning (RL), which often fails to converge when processing high-dimensional visual inputs \cite{xing_bootstrapping_2024}, IL provides stable training by leveraging expert demonstrations \cite{Zare2024TCYB}. To bridge the gap between local sensing and global awareness, privileged learning has emerged as a powerful paradigm \cite{chen_learning_2020}. By training a ``student" policy to mimic a ``teacher" that has access to ground-truth environmental data (privileged information) during training, the student can learn to implicitly reconstruct hidden features from raw sensor histories \cite{Lee2020SR,miki_learning_2022}. However, traditional two-stage privileged learning—where the teacher is first trained via RL—suffers from low sample efficiency and unpredictable expert performance \cite{wang_synergistic_2024}.
However, its advantages remain largely unexploited in vision-based autonomous UAV flight. Furthermore, vanilla privileged learning typically involves a two-stage learning process \cite{chen_learning_2020,wang_synergistic_2024}. In the first stage, the teacher policy is learned using an RL algorithm, while in the second stage, the student policy is trained to mimic the teacher. It often requires repeated trial and error, as the performance of the learned teacher policy is not guaranteed and is highly uncertain. Consequently, the two-stage learning process in vanilla privileged suffers from low sample efficiency and Unstable training,  resulting in high computational demands and prolonged training times.
 
In this paper, we propose PILOT, a constraint-aware,  \textbf{\underline{P}}rivileged \textbf{\underline{I}}mitation \textbf{\underline{L}}earning algorithm for vision-based end-to-end UAV m\textbf{\underline{O}}\textbf{\underline{T}}ion planning in diverse cluttered environments. Unlike traditional two-stage privileged learning \cite{xing_bootstrapping_2024, wang_synergistic_2024}, PILOT adopts a one-stage learning paradigm, employing an MPC expert with access to full privileged information—including complete vehicle states and precise obstacle locations—to provide high-quality trajectory supervision under the stated model and optimization assumptions. A trajectory parameterization layer is integrated into the End-to-End (E2E) architecture of the learned planner, enabling PILOT to regress control points rather than raw control inputs. This design ensures an improved adherence to UAV dynamic constraints, particularly for fixed-wing UAVs. To address partial observability, a Temporal Convolutional Network (TCN) \cite{Lea2017CVPR} is incorporated to implicitly infer latent features from historical visual observations, thereby reconstructing the privileged spatial context necessary for navigation beyond the sensor’s immediate field of view (FOV). The PILOT framework is evaluated via Monte Carlo simulations on both quadrotor and fixed-wing platforms to demonstrate its cross-platform generality. Ablation studies are performed to verify the efficacy of both the trajectory parameterization and TCN modules. Sim-to-real transfer is demonstrated through zero-shot deployment on a custom-built quadrotor in both indoor and outdoor settings. The learned policy achieves comparable performance to the privileged expert while reducing computational overhead by more than 80\% across various settings.

\subsection{Related Works}
\label{sec:related_works}
Vision-based planning is widely explored for its efficiency and flexibility in unstructured, congested environments. Conventional frameworks adopt a modular approach, partitioning tasks into independent perception, prediction, and planning sub-tasks \cite{Frazzoli2002JGCD, Guo2020IJRR, Tang2024TAES, lu_autonomous_2025}. Though sophisticated methods—including vector field \cite{Guan2020TAES, Marchidan2020JGCD}, sampling-based \cite{Frazzoli2002JGCD}, and optimization-based designs \cite{lu_autonomous_2025, Zhu2025T-Mech}—have been developed, these methods rely on privileged environmental information.  However, assuming full access to privileged data is impractical for real-world integration. Many designs \cite{Wilhelm2019JGCD, Marchidan2020JGCD, Zhang2022JGCD, Zhu2025T-Mech} simplify obstacles as circles with known parameters—an assumption that often fails or leads to overly conservative performance in complex scenes. Furthermore, it is difficult to precisely determine obstacle geometry with onboard perception. While representing environments via discrete cells or voxel grids can mitigate those issues, it significantly increases the computational costs \cite{Tordesillas2021TRO, Toumieh2024TRO}. 

While vision-based modular methods offer interpretability, their extended workflows demand significant expert knowledge and intricate design. This multi-stage pipeline is prone to cumulative errors and latencies resulting from sequential, computationally intensive tasks. To address these drawbacks, learning-based methods have emerged as a compelling alternative, providing E2E solutions that circumvent the complexities of conventional modular architectures \cite{loquercio_dronet_2018}.  Existing research employs RL or IL to develop E2E policies \cite{loquercio_dronet_2018,Schilling2019RAL}. In RL, optimal policies are trained offline via trial and error, which requires an immense volume of samples. Conversely, IL demonstrates superior sample efficiency when processing high-dimensional visual inputs, making it a focal point for contemporary robot learning research \cite{pinto_asymmetric_2018,ZYXiao_2023_IROS,miki_learning_2022}.

IL enables policies to be learned from expert demonstrations, such as optimization-based planners \cite{tordesillas_deep-panther_2023, zhang_learning_2016} or human pilots \cite{ross_learning_2013}, though performance is bounded by the expert's competence. To transcend this, privileged imitation learning \cite{chen_learning_2020} employs a privileged expert to guide a student policy via knowledge distillation. However, direct control action generation \cite{fu_learning_2023, song_learning_2023, loquercio_dronet_2018} lacks motion smoothness, compromising reliability during scene transfer. Alternatively, methods using predefined motion primitives \cite{nguyen_motion_2022, lu_lpnet_2023} are limited by discrete partitioning, leading to suboptimal performance.  In contrast, our framework mimics a privileged expert based on a set of control points for parameterized trajectories rather than single-step actions. This trajectory parameterization approach provides a structured output to improves motion smoothness with the adherence to dynamic limits.

Unlike decoupled approaches, PILOT unifies privileged imitation learning, temporal perception encoding, and optimization-based planning around a shared Bézier trajectory representation, enabling direct trajectory-level knowledge transfer and a compact policy deployable under partial observability.

\subsection{Contributions}
First, we propose a constraint-aware privileged imitation learning framework for vision-driven autonomous flight. Unlike conventional behavior cloning or two-stage teacher-student paradigms, the framework distills a computationally intensive privileged expert into a deployable student policy via a single-stage training objective. By minimizing a dual-objective loss that couples trajectory imitation with constraint-related regularization, the policy promotes adherence to dynamic and safety requirements even in the presence of imitation errors.

Second, we propose a spatiotemporal perception fusion module to address partial observability in vision-based navigation. Unlike single-frame policies or explicit mapping pipelines, the module leverages a TCN to causally integrate finite histories of depth images and odometry. It enables the planner to infer task-relevant latent context, compensating for limited fields of view while retaining a lightweight representation suitable for real-time onboard inference.

Third, we integrate B\'ezier trajectory parameterization into the end-to-end planner. Rather than predicting pointwise control commands, the network regresses a set of control points used to generate a finite-horizon trajectory. This formulation provides a $C^2$-continuous trajectory representation, enabling direct trajectory-level supervision, analytic evaluation of continuity and dynamic penalties during training, and platform-specific reference extraction without post hoc fitting.

The rest of the paper is organized as follows. Section \ref{sec:Prob} introduces preliminaries for the algorithm design and formulates the learning-based planning problem. In Section \ref{sec:method}, the PILOT framework is presented. Numerical simulations are provided in Section \ref{sec:sim}, followed by real-world experimental results in Section~\ref{sec:exp}. Finally, Section~\ref{sec:conclusion} concludes the paper with summary remarks.

\section{\textsc{PROBLEM FORMULATION}\label{sec:Prob}}

In this section, the mathematical foundation is established by defining the system dynamics and the privileged optimal control problem, followed by a formalization of partial observability inherent in vision-based flight.

\subsection{System Dynamics}
The proposed PILOT framework is designed to be platform-agnostic, facilitating deployment across heterogeneous aerial platforms. Hence, a general nonlinear discrete-time model is considered.
\begin{equation}\label{eq:UAVmodel}
    \boldsymbol{x}_{t+1} = \boldsymbol{f}\left(\boldsymbol{x}_t,\boldsymbol{u}_t\right)
\end{equation}
where $\boldsymbol{x}_{t}\in\mathbb{R}^{n_{x}}$ is the state vector of a UAV at the time step $t$,   $\boldsymbol{u}_{t}\in\mathbb{R}^{n_{u}}$ denotes the control input, and $\boldsymbol{f}:\mathbb{R}^{n_{x}}\times \mathbb{R}^{n_{u}}\to\mathbb{R}^{n_{x}}$ represents the general dynamics. 
Following established literature \cite{Schmitt2014JGCD, Marchidan2020JGCD, Adhikari2020JGCD}, a UAV is assumed to be equipped with an inner-loop attitude controller, which allows us to focus on translational maneuvers. Accordingly, the state vector is  $\boldsymbol{x}_{t}=[\boldsymbol{p}_t^\top, \boldsymbol{v}_t^\top]^\top$, where $\boldsymbol{p}_t, \boldsymbol{v}_t \in \mathbb{R}^{n}$ denote the position and velocity in the inertial frame ($n \in \{2, 3\}$). In this study, acceleration is chosen as the input $\boldsymbol{u}_{t}$, which is readily transformable into attitude commands—represented via $SO(3)$ or quaternions—to be tracked by the inner-loop autopilot. A UAV operates under state and input constraints such that $\boldsymbol{x}_{t}\in\mathcal{C}_{\boldsymbol{x}}$ and $\boldsymbol{u}_{t}\in\mathcal{C}_{\boldsymbol{u}}$, where $\mathcal{C}_{\boldsymbol{x}}$ and $\mathcal{C}_{\boldsymbol{u}}$ are assumed to be convex. 

\subsection{Privileged Constrained Optimal Control Problem}\label{subsec:PCOCP}
One central component of our design is the ``privileged" expert that provides the supervision signal for the learning process. The privileged state $\boldsymbol{s}_{p} \in \mathcal{S}_{p}$  is defined as an augmented vector with $\boldsymbol{s}_{p} = \left\{\boldsymbol{x}, \boldsymbol{x}_{env}\right\}$, where $\boldsymbol{x} \in \mathbb{R}^{n_x}$ is the UAV's system state in \eqref{eq:UAVmodel}, and $\boldsymbol{x}_{env}$ is the environmental information, such as global occupancy maps, obstacle locations and geometries. Note that $\mathbf{s}_{p}$ provides an omniscient view of the navigation manifold. The privileged planning task is thus formulated as a receding-horizon Constrained Optimal Control Problem (OCP). At each time step $t$, the expert policy $\pi^*$ generates an optimal control sequence $\boldsymbol{u}^*_{t:t+H}$ by solving
\begin{subequations}\label{eq:OCP}
\begin{align}
    \min_{\boldsymbol{u}_{t:t+H}} \quad & \sum_{k=t}^{t+H} \ell\left(\boldsymbol{x}_k, \boldsymbol{u}_k, \boldsymbol{x}_{g}\right) \\
    \text{s.t. } \quad & \boldsymbol{x}_{k+1} = \boldsymbol{f}(\boldsymbol{x}_k, \boldsymbol{u}_k), \label{eq:dyn_cons} \\
    & \boldsymbol{x}_k \in \mathcal{X}_{free}(\boldsymbol{s}_{p}), \label{eq:obs_cons} \\
    & \boldsymbol{x}_k \in \mathcal{C}_{\boldsymbol{x}}, \quad \boldsymbol{u}_k \in \mathcal{C}_{\boldsymbol{u}}, \label{eq:limits_cons}
\end{align}
\end{subequations}
where $\ell$ is a cost functional penalizing deviation from the goal and control effort. This OCP acts as the framework's oracle, yielding the optimal state-action pairs that constitute the target manifold for training our E2E policy.

\subsection{Partial Observability}\label{subsec:PO}
While the expert policy $\pi^*$ operates on the omniscient privileged state $\boldsymbol{s}_{p}$, real-world deployment is restricted to a partial observation $\boldsymbol{o}_t = \gamma(\boldsymbol{s}_{p,t})$, where $\boldsymbol{o}_t$ represents a high-dimensional sensor stream, such as a FOV-constrained depth image or a local point cloud. 
The mapping $\gamma: \mathcal{S}_p \to \mathcal{O}$ models sensory limitations—including finite range, restricted FOV, and measurement noise—thereby inducing a Partially Observable Markov Decision Process (POMDP). Since an instantaneous observation $\boldsymbol{o}_t$ is insufficient to uniquely reconstruct the collision-free space $\mathcal{X}_{\mathrm{free}}$, we use $\boldsymbol{o}_{t-k:t} = \{\boldsymbol{o}_{t-k+1}, \dots, \boldsymbol{o}_t\}$ over a lookback horizon $k$. Consequently, the planning task reduces to learning a parameterized policy $\pi_{\boldsymbol{\theta}}(\boldsymbol{o}_{t-k:t}, \boldsymbol{x}_{g})$ that maps the observation history and goal state $\boldsymbol{x}_g$ to the optimal control manifold. The policy $\pi_{\boldsymbol{\theta}}$ must satisfy the following requirements.
\begin{itemize}
    \item \textbf{Collision-free:} $\boldsymbol{x}_t \in \mathcal{X}_{free}(\boldsymbol{s}_{p})$, $\forall t\in\left[t_0\text{, }t_f\right]$, where $t_f$ is the total time to achieve the navigation task;
    \item \textbf{Dynamically feasible:} $\boldsymbol{x}_t\in\mathcal{C}_{\boldsymbol{x}}$ and $\boldsymbol{u}_t\in\mathcal{C}_{\boldsymbol{u}}$, $\forall t\in\left[t_0\text{, }t_f\right]$;
    \item \textbf{Computationally efficient:} $\pi_{\theta}$ is suitable for real-time execution on resource-constrained hardware. 
\end{itemize} 
By leveraging temporal correlations, $\pi_{\boldsymbol{\theta}}$ aims to recover the latent environmental features encoded in $\boldsymbol{s}_p$, effectively approximating the expert’s performance under sensory and computational constraints. The dynamic feasibility and collision-avoidance criteria defined above specify operational targets rather than formal guarantees. PILOT incorporates soft penalties for velocity, acceleration, continuity, and obstacle clearance to encourage constraint-consistent trajectories. However, these soft-penalty losses do not guarantee constraint satisfaction for out-of-distribution observations or under perception noise.

\section{\textsc{PILOT Framework}}
\label{sec:method}
\begin{figure*}[tbhp]
        \centering
        \includegraphics[width=0.975\linewidth]{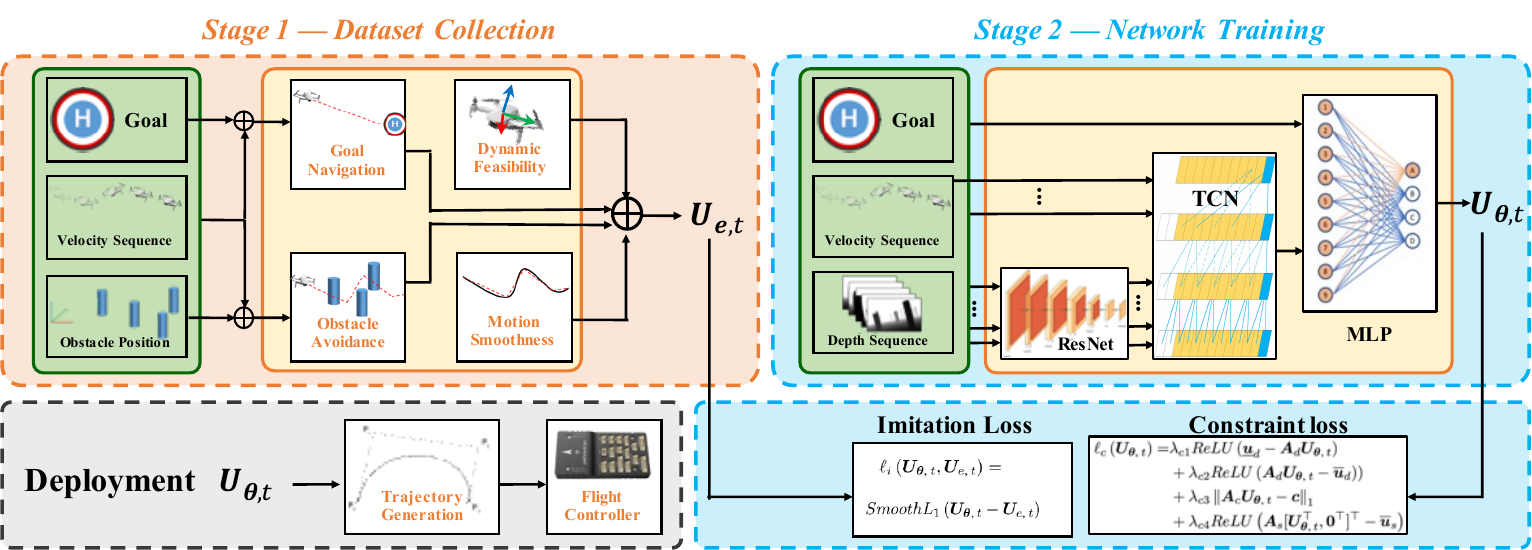}
        \caption{The proposed PILOT framework. The network policy takes a sequence of depth images, historical measurements of velocities, and the goal state as inputs, while the MPC expert has access to privileged information, \emph{e.g.} locations of surrounding obstacles.}
        \label{fig:Arch}
    \end{figure*}

The PILOT framework enables vision-driven E2E motion planning using a privileged optimization-based expert as the training  mentor (Fig.~\ref{fig:Arch}). Unlike behavior cloning, PILOT employs a guided learning paradigm with the student policy optimized against a dual-objective that balances expert trajectory imitation with explicit penalties for predicted system constraint violations.
    
\subsection{Trajectory Representation and Mapping}\label{subsec:trajRep&Map}
Rather than imitating low-level control commands \cite{wang_synergistic_2024, xing_bootstrapping_2024}, PILOT distills globally informed planning behaviors into a deployable policy operating under partial onboard observations. To this end, both the privileged expert and the student policy share an identical Bézier trajectory representation. The expert generates trajectory control points using full environmental information, while the student learns to predict the same control points from onboard perception. The trajectory parameterization enables the derivation of continuous trajectories from discrete control points, endowing the policy with multi-step motion awareness and proactive collision risk assessment.

Let $\boldsymbol{U} = \{\boldsymbol{P}_0, \boldsymbol{P}_1, \dots, \boldsymbol{P}_m\}$ be a set of $m+1$ control points. A B\'ezier curve $\boldsymbol{b}(t)$ is defined as
\begin{equation} \label{eq:BezierCurve}
    \boldsymbol{b}(t) = \sum_{i=0}^{m} \boldsymbol{P}_i b_{i,m}(t), \quad t \in [0, T]
\end{equation}
where $T > 0$ is the duration of the planning horizon, and $b_{i,m}(t)$ are the Bernstein basis polynomials of degree $m$.
\begin{equation}
    \label{pos_compute}
    b_{i,m}(t) = \binom{m}{i} \left(1 - \frac{t}{T}\right)^{m-i} \left(\frac{t}{T}\right)^i, \quad t \in [0, T]
\end{equation}
where $\binom{m}{i}$ represents the binomial coefficient. 

The derivatives of the Bézier curve \eqref{eq:BezierCurve} can be expressed using the same set of control points as below.
\begin{equation}
    \label{vel_compute}
    \dot{\boldsymbol{b}}(t) = \sum_{i=0}^{m-1} \frac{m}{T}\left(\boldsymbol{P}_{i+1} - \boldsymbol{P}_i\right) b_{i,m-1}(t)
\end{equation}
\begin{equation}
    \label{control_input_compute}
    \ddot{\boldsymbol{b}}(t) = \sum_{i=0}^{m-2} \frac{m(m-1)}{T^2}\left(\boldsymbol{P}_{i+2} - 2\boldsymbol{P}_{i+1} + \boldsymbol{P}_i\right) b_{i,m-2}(t) 
\end{equation}
This formulation preserves computational efficiency on resource-constrained hardware. Here, the subscript $t+h$ indicates a predicted quantity $h$ steps ahead of current time $t$, for $h \in \{1, \dots, H\}$ with prediction horizon $H$. Extended trajectories are represented as piecewise Bézier curves formed by concatenating $M$ segments.

B\'ezier curves possess the convex hull property, ensuring that the curve $\boldsymbol{b}(t)$ lies strictly within the convex hull of its control points $\boldsymbol{U}$. Because the derivatives in \eqref{vel_compute} and \eqref{control_input_compute} are themselves B\'ezier curves formed by the forward differences of $\boldsymbol{U}$, they likewise inherit this property. In PILOT, representing the finite-horizon trajectory via multiple B\'ezier segments localizes the convex-hull bounds to individual segments, reducing the conservatism of a single B\'ezier curve. While the geometric convex-hull property itself is exact, practical collision avoidance also depends on obstacle representations, local linearization, and slack variables; thus, it is not interpreted as an unconditional safety guarantee. For scenarios requiring topological or route-level decisions (e.g., sharp turns or narrow passages), PILOT functions as a local planner that can be guided by an upper-level planner providing intermediate waypoints, reference paths, or safe corridors.

\subsection{Expert Policy via Constrained Optimization}\label{subsec:MPC}
The expert policy is realized through Model Predictive Control (MPC), which serves as the numerical implementation of the Privileged Constrained OCP defined in Section \ref{sec:Prob}-\ref{subsec:PCOCP}. By adopting the B\'ezier parameterization introduced in Section \ref{sec:method}-\ref{subsec:trajRep&Map}, the OCP is transformed into a structured Quadratic Programming (QP) problem.
\begin{subequations}\label{eq:mpc_qp}
\begin{align}
\min_{\boldsymbol{U}_{t},\,\boldsymbol{\Sigma}_{t}} \quad & J_{g} + J_{u} + J_{s}  \label{eq:mpc_qp_obj}\\
\text{s.t.}\quad 
& \boldsymbol{x}_{t+i\left| t\right.}\in\mathcal{C}_{\boldsymbol{x}} \text{ } \& \text{ } \boldsymbol{u}_{t+i\left| t\right.}\in\mathcal{C}_{\boldsymbol{u}}\text{ }  \label{eq:mpc_qp_dyn}\\
& \boldsymbol{x}_{t+i\left| t\right.} \in \mathcal{X}_{free}(\boldsymbol{s}_{p}) \text{, }\forall i\in\left\{1\text{, }2\text{, }\ldots\text{, }H\right\} \label{eq:mpc_qp_safe}\\
& \mathcal{C}_{\boldsymbol{c}}\left(\boldsymbol{U}_{t}, \boldsymbol{c}\right) =\boldsymbol{0} \label{eq:mpc_qp_con}\\
&\mathcal{C}_{\boldsymbol{\Sigma}}\left(\boldsymbol{\Sigma}_{t}\right) \le\ \boldsymbol{0} \label{eq:mpc_qp_slack}
\end{align}
\end{subequations}
where  $\boldsymbol{U}_{t}$ is the stacked vector of control points defining the trajectory manifold $\boldsymbol{b}(t)$ in \eqref{eq:BezierCurve}, and $\boldsymbol{\Sigma}_{t}$ is the vector of slack variables. The terms $J_g$, $J_u$, and $J_s$ represent the navigation cost, control input penalty, and constraint violation cost, respectively. The subscript ${t+i|t}$ denotes a predicted value at time $t+i$ given information available at time $t$, and $H$ defines the total prediction horizon. Finally, $\mathcal{C}_{\boldsymbol{c}}(\boldsymbol{U}_{t}, \boldsymbol{c})$ and $\mathcal{C}_{\boldsymbol{\Sigma}}(\boldsymbol{\Sigma}_{t})$ denote the constraint functions for the control points and slack variables.

\subsubsection{Cost Function formulation} 
The objective function \eqref{eq:mpc_qp_obj} is designed to balance navigation efficiency, motion smoothness, and constraint feasibility. To ensure flight convergence toward the target, the navigation cost $J_g$ is defined as the cumulative squared Euclidean distance between the predicted positions $\boldsymbol{p}_{t+i|t}$ and the goal position $\boldsymbol{p}_{g}$, so
\begin{equation}
    J_{g} = \sum_{i = 1}^H \lambda_{g}\left\Vert\boldsymbol{p}_{t+i\left| t\right.} - \boldsymbol{p}_{g}\right\Vert_2^2
    \end{equation}
where $\left\Vert\cdot\right\Vert_2$ denotes the Eucildean norm, and $\lambda_{g}>0$ is the associated weighting scalar.

To penalize aggressive maneuvers and encourage trajectory smoothness, the control cost $J_u$ is defined as
\begin{equation}
    J_{u} = \sum_{i = 1}^H(\lambda_{u}\left\Vert\boldsymbol{u}_{t+i\left| t\right.}\right\Vert_2^2)
    \end{equation}
where  $\lambda_{u} > 0$ is the penalty weight for the control effort.

Furthermore, the constraint violation cost $J_s$ handles obstacle avoidance while maintaining numerical stability. Introducing slack variables $\boldsymbol{\Sigma}_{t} = [\boldsymbol{\sigma}_{t+1|t}^\top, \dots, \boldsymbol{\sigma}_{t+H|t}^\top]^\top \in \mathbb{R}^{N_pH}$, with $N_p$ denoting obstacle-surface points within the privileged perception range, relaxes collision constraints into soft penalties. This relaxation enhances numerical feasibility near clearance boundaries or in cluttered spaces. However, since positive slack values reduce clearance margins, QP feasibility alone does not certify a collision-free path. The constraint penalty is 
\begin{equation}
    J_{s} = \sum_{i = 1}^H(\lambda_{s\text{, }1}\boldsymbol{1}_{N_p}^\top\boldsymbol{\sigma}_{t+i\left| t\right.}+\lambda_{s\text{, }2} \left\Vert\boldsymbol{\sigma}_{t+i\left| t\right.}\right\Vert_2^2)
\end{equation}
where $\lambda_{s\text{, }1} > 0$ and $\lambda_{s\text{, }2} > 0$ are constant weights, and $\boldsymbol{1}_{N_p} \in \mathbb{R}^{N_p}$ is a column vector of ones. This formulation provides the optimizer greater flexibility while penalizing the magnitude of nominal-clearance violations.

\subsubsection{Constraint Mapping via Convex Hull}
State and control constraints in \eqref{eq:mpc_qp_dyn} restrict the trajectory to the UAV's physical limits, given by $\mathcal{C}_{\boldsymbol{x}} = \{ \boldsymbol{x} \in \mathbb{R}^n \mid \underline{\boldsymbol{x}} \leq \boldsymbol{x} \leq \overline{\boldsymbol{x}} \}$ and $\mathcal{C}_{\boldsymbol{u}} = \{ \boldsymbol{u} \in \mathbb{R}^n \mid \underline{\boldsymbol{u}} \leq \boldsymbol{u} \leq \overline{\boldsymbol{u}} \}$. By the convex hull property of B\'ezier curves (Section \ref{sec:method}-\ref{subsec:trajRep&Map}), these constraints translate into bounded inequalities on the control points.
\begin{equation}\label{eq:dynamic_constraints}
    \underline{\boldsymbol{u}}_{d} \leq \boldsymbol{A}_{d} \boldsymbol{U}_{t} \leq \overline{\boldsymbol{u}}_{d}
\end{equation}
where $\boldsymbol{A}_{d}$ is a constraint matrix mapping the stacked control points to the corresponding derivatives.
 
To enforce $C^2$ continuity across the junctions of $M$ concatenated B\'ezier segments, the equality constraint $\mathcal{C}_{\boldsymbol{c}}(\boldsymbol{U}_{t}, \boldsymbol{c}) = \boldsymbol{0}$ is formulated as
\begin{equation}
    \mathcal{C}_{\boldsymbol{c}}\left(\boldsymbol{U}_{t}, \boldsymbol{c}\right)=\boldsymbol{A}_{c} \boldsymbol{U}_{t} -\boldsymbol{c}= \boldsymbol{0}
\end{equation}
where $\boldsymbol{A}_{c}$ encodes boundary conditions linking final control points of segment $j$ to initial points of segment $j+1$.

\subsubsection{Privileged Obstacle Avoidance}
To facilitate collision avoidance in cluttered environments, the MPC expert maintains a nominal clearance from obstacles. Privileged obstacle information is represented by a dense point cloud $\mathcal{P}_{o}=\{\boldsymbol{p}_{o,j}\}_{j=1}^{N_p}$ sampled across occupied surfaces within the perception range. This point-cloud representation unifies diverse geometries—sampling lateral/end surfaces of upright and rigidly transformed tilted cylinders, exterior faces of cuboids, and frame surfaces of gate obstacles to preserve free openings. The collision-avoidance condition is thus given by
\begin{equation} \label{eq:nonlinCollAvoidaceConstraint}
\left\|\boldsymbol{p}_{t+i|t} - \boldsymbol{p}_{o,j}\right\|_2 \geq d_{s} - \sigma_{t+i|t,j}\text{, } \forall j \in \{1, \dots, N_p\}
\end{equation}
where $\boldsymbol{p}_{o,j}$ is the $j$-th obstacle-surface point, $d_s$ is the required clearance incorporating vehicle safety margins, $N_p$ is the count of evaluated surface points in the privileged range, and $\sigma_{t+i|t,j}$ is the slack variable. This point-cloud representation provides a discrete surface approximation whose resolution depends on the sampling interval.

For efficient online implementation, the nonlinear constraint \eqref{eq:nonlinCollAvoidaceConstraint} is linearized via a first-order Taylor expansion, yielding the following linear inequality.
\begin{equation}
    \boldsymbol{A}_{s}[\boldsymbol{U}_t^\top, \boldsymbol{\Sigma}_t^\top]^\top \leq \overline{\boldsymbol{u}}_{s}
    \end{equation}
where $\boldsymbol{A}_{s}$ is a constant matrix and $\overline{\boldsymbol{u}}_{s}$ is an upper bound vector. They are constructed from the privileged obstacle-surface point cloud $\mathcal{P}_o$ in \eqref{eq:nonlinCollAvoidaceConstraint}. The slack variables satisfy non-negativity constraints, so $\mathcal{C}_{\boldsymbol{\Sigma}}\left(\boldsymbol{\Sigma}_{t}\right)$ is given by
    \begin{equation}
\mathcal{C}_{\boldsymbol{\Sigma}}\left(\boldsymbol{\Sigma}_{t}\right) =-\boldsymbol{\Sigma}_t\leq\boldsymbol{0}
    \end{equation}
The QP problem \eqref{eq:mpc_qp} is solved via OSQP \cite{Stellato_OSQP_2018}, generating optimal control points $\boldsymbol{U}_{e,t}$ for the E2E planner to mimic.

\subsection{Temporal Encoding for Latent State Inference}
The E2E trajectory planner maps high-dimensional visual observations and platform dynamics into feasible trajectories by imitating expert decision-making. As illustrated in Fig. \ref{fig:Arch}, the architecture comprises three primary components: an image processing module for spatial feature extraction, a spatio-temporal fusion module for sequential dependency modeling, and a trajectory planning module for control point generation. Due to the limited FOV of the onboard camera, a UAV operates under partial observability, making an instantaneous observation $\boldsymbol{o}_t$ insufficient to capture the complete environment state. To address this limitation, the architecture employs temporal encoding to infer a latent state from historical context.

To extract structural cues from raw depth data, a ResNet-18 backbone \cite{He2016CVPR} is used, which processes a sequence of depth images $\boldsymbol{o}_{t-k:t}$ with $k$ denoting the look-back window. Each residual building block is defined by 
\begin{equation} \label{eq:ResNetBlock}
\boldsymbol{y} = \boldsymbol{f}_{NN}\left(\boldsymbol{z},{\boldsymbol{W}_1, \boldsymbol{W}_2}\right) + \boldsymbol{z}
\end{equation}
where $\boldsymbol{z}$ and $\boldsymbol{y}$ are the block input and output, respectively, and $\boldsymbol{f}_{NN}(\cdot)$ represents two convolutional layers with weights $\boldsymbol{W}_1$ and $\boldsymbol{W}_2$. The encoder maps each observation $\boldsymbol{o}_{t}$ to a compact spatial feature $\boldsymbol{F}_t \in \mathbb{R}$.

    \begin{figure}[tbph]
        \centering
      \includegraphics[width=0.9\linewidth]{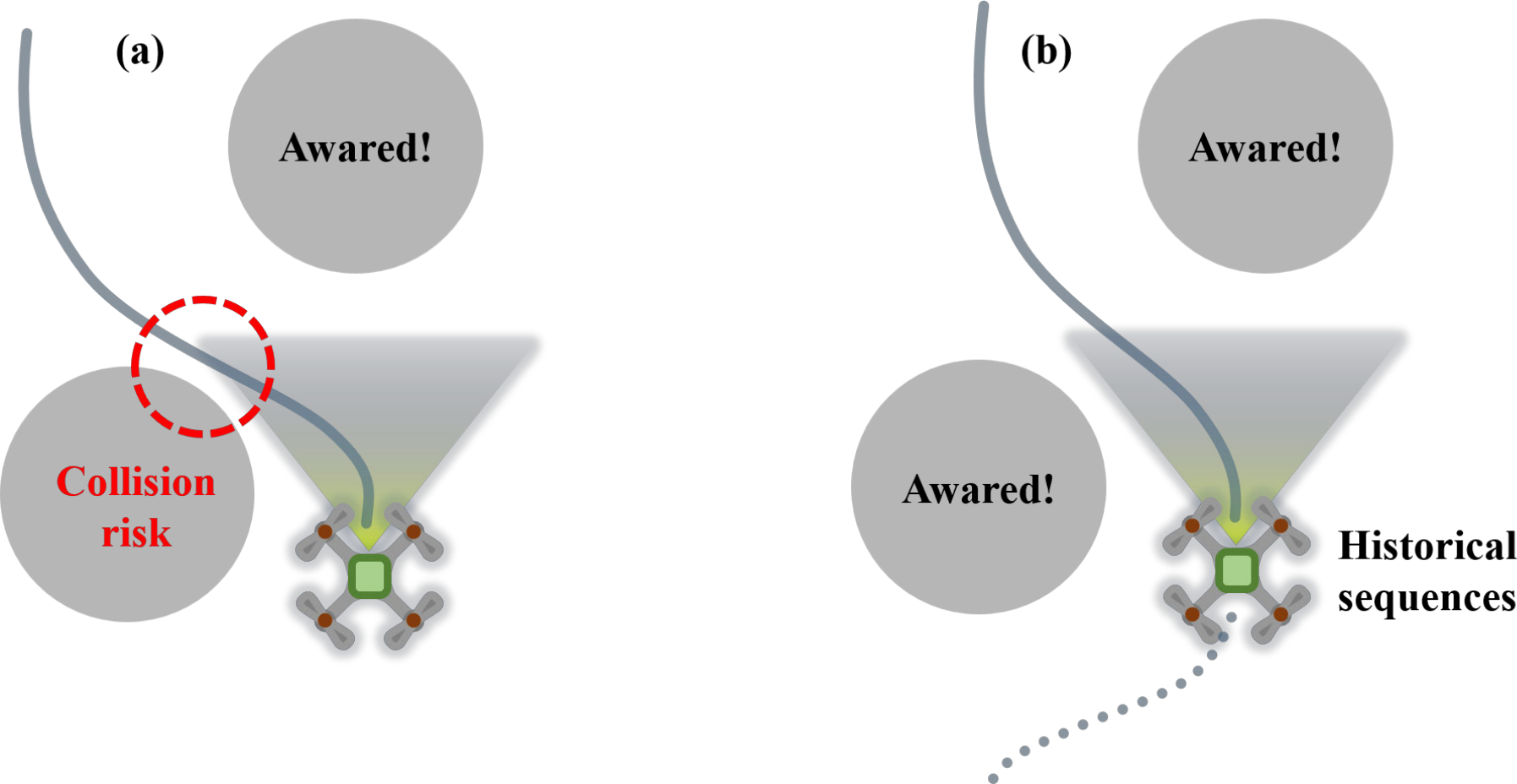}
        \caption{Perception modules. (a) Instantaneous observations: Limited FOV risks collisions in unobserved regions. (b) Temporal encoding: Spatiotemporal feature integration improves scene understanding to enable collision-free navigation.}
        \label{fig:obs aware}
    \end{figure}

While the MPC expert uses privileged environmental information, the E2E planner reasons about the 3D scene from historical observation sequences. A TCN is employed to infer task-relevant spatiotemporal context from the finite observation window. Compared to recurrent architectures, TCNs offer improved gradient stability and parallelized computation. The spatial feature $\boldsymbol{F}_{t}$ is concatenated with the current velocity $\boldsymbol{v}_{t}$ to form an augmented feature vector $\bar{\boldsymbol{F}}_{t} = [\boldsymbol{F}_{t}^\top, \boldsymbol{v}_{t}^\top]^\top$, forming the input matrix $\bar{\boldsymbol{F}}_{t-k:t}$ over the temporal window. A three-layer TCN with dilated causal convolutions (dilation factors $[1, 2, 4]$) expands the receptive field to capture multiscale dependencies while strictly preserving causality ($\tau \leq t$). As illustrated in Fig. \ref{fig:obs aware}, planners based on instantaneous observations face higher collision risks in constrained geometries due to limited FOV. In contrast, the TCN-based approach integrates recent observations to reduce spatial ambiguity and retain short-term obstacle context.

The TCN outputs a latent spatiotemporal matrix $\boldsymbol{H}$, which is compressed into a global embedding $\boldsymbol{h}_o$ and concatenated with target goal features for downstream planning. This embedding provides temporal consistency to support navigation under partial observability.

\subsection{Constraint-Aware Policy Distillation} \label{subsec:CAPolicy}
Our constraint-aware PIL leverages B\'ezier curves, enabling the student policy to distill full teacher-planned trajectories rather than single-step actions. Furthermore, it incorporates explicit constraint penalties directly into the imitation learning objective.

\subsubsection{Trajectory Planning Module}
The trajectory planning module transforms the latent environmental embedding $\boldsymbol{h}_o$ and goal state $\boldsymbol{x}_g$ into actionable control points. To facilitate directional reasoning, the goal position $\boldsymbol{p}_g$ (extracted from $\boldsymbol{x}_g$) is pre-processed into a goal direction vector $\boldsymbol{g}_{d} \in \mathbb{R}^3$.
    \begin{equation}
        \boldsymbol{g}_{d} = \begin{cases}
        \frac{\boldsymbol{p}_{t} - \boldsymbol{p}_{g}}{\left\Vert\boldsymbol{p}_{t} - \boldsymbol{p}_{g}\right\Vert_2}, & \text{if } \left\Vert\boldsymbol{p}_{t} - \boldsymbol{p}_{g}\right\Vert_2 \geq 1 \\
        \boldsymbol{p}_{t} - \boldsymbol{p}_{g}, & \text{otherwise}
        \end{cases}
    \end{equation}
This module is implemented as a $2$-layer MLP with 256 hidden units and ReLU activations. It generates a set of control points $\boldsymbol{U}_{\boldsymbol{\theta}, t}$ characterizing a piecewise Bézier trajectory. This parameterization allows the distillation process to evaluate system, continuity, and obstacle-clearance residuals directly from the student output, thereby encouraging smooth and constraint-consistent trajectories.

\subsubsection{Privileged Imitation Learning Objective}
The distillation process is governed by two synergistic objectives: a primary imitation loss to match the expert's behavior and a constraint-aware loss to restrict the output space. Given a dataset $\mathcal{D} = \{(\boldsymbol{o}_{t-k:t}, \boldsymbol{x}_{g}, \boldsymbol{U}_{e,t})\}_{t=1}^N$ collected from the MPC expert $\pi_e$, the optimal parameters $\boldsymbol{\theta}^*$ are found by minimizing
\begin{equation} \label{eq:PIL_Objective}
\begin{aligned}
    \boldsymbol{\theta}^*=\arg\min_{\boldsymbol{\theta}} \; \mathbb{E}_{\left(\boldsymbol{o}_{t-k:t}\text{, }\boldsymbol{x}_{g}\text{, }\boldsymbol{U}_{e\text{, }t}\right) \sim \mathcal{D}} &
     \left[ \ell_{i}\left(\boldsymbol{U}_{\boldsymbol{\theta}\text{, }t},\boldsymbol{U}_{e\text{, }t}\right)\right.\\
     &\left.+\ell_{c}\left(\boldsymbol{U}_{\boldsymbol{\theta}\text{, }t}\right)\right]
\end{aligned}
\end{equation}
where $\boldsymbol{U}_{\boldsymbol{\theta}\text{, }t}$ is the control point output of the student policy $\pi_{\boldsymbol{\theta}}$, $\ell_i(\cdot)$ is a similarity metric measuring the distance between the learned policy $\pi_{\boldsymbol{\theta}}$ and its demonstrator $\pi_{e}$, $\ell_{c}$ is a loss function related to constraints.
\begin{algorithm}[tbp]
    \caption{\textbf{PILOT Algorithm}}
    \label{pilot_training}
    \begin{algorithmic}
        \REQUIRE{Dataset $\mathcal{D}$, expert policy $\pi_{e}$}, environment $\mathcal{E}$, learning rate $\eta$
        \ENSURE{Student policy $\pi_{\boldsymbol{\theta}}$}
        \STATE{Initialize $\mathcal{D}\xleftarrow{}\emptyset$}
        \FOR{$ i=1, 2,\cdots, I$}
            \STATE $\boldsymbol{o}_i\text{, }\mathcal{X}_{o}\text{, }\boldsymbol{x}_g \gets \text{ResetEnvironment}(\mathcal{E})$
            \FOR{$t=1,2,\cdots,T$}
                \STATE $\boldsymbol{o}_{t-k:t}\text{, }\boldsymbol{x}_g \gets \text{GetDepthImageSequenceAndState}(\mathcal{E})$
                
                \STATE $\boldsymbol{U}_{e,t} \gets \pi_{e}(\boldsymbol{s}_p\text{, }\boldsymbol{x}_g\text{, }\mathcal{X}_{o})$ \ // \ \text{Eq.}\eqref{eq:mpc_qp}

                \STATE $\boldsymbol{u}_t \gets \text{ComputeControlInput}(\boldsymbol{U}_{e,t})$
                \COMMENT{Eqs.~\eqref{pos_compute}--\eqref{control_input_compute}}

                \STATE $\boldsymbol{x}_{t+1}\text{,}\boldsymbol{o}_{t+1} \gets \text{EnvironmentStep}(\boldsymbol{u}_t)$
                \STATE $\mathcal{D}\gets\mathcal{D}\cup\left(\boldsymbol{o}_{t-k:t}\text{, }\boldsymbol{x}_{g}\text{, }\boldsymbol{u}_{t-1}\text{, }\boldsymbol{U}_{e,t}\right)$
                
                \STATE $t \gets t+1$
            \ENDFOR
        \ENDFOR
        \STATE Initialize the student policy $\pi_{\boldsymbol{\theta}}$
        \FOR{$ e=1, 2,\cdots, E$}
            \STATE $\mathcal{B} \gets \text{SampleBatch}(\mathcal{D})$
            \FOR{each $(\boldsymbol{x}_t\text{, } \boldsymbol{o}_{t-k:t}\text{, }\boldsymbol{p}_{g}\text{, }\boldsymbol{u}_{t-1}\text{, }\boldsymbol{U}_{e,t}) \in \mathcal{B}$}
                \STATE ${\boldsymbol{U}}_{\boldsymbol{\theta}\text{, }t} \gets \pi_{\boldsymbol{\theta}}(\boldsymbol{o}_{t-k:t}\text{, }\boldsymbol{x}_{g})$
                \STATE $\mathcal{L} \gets \mathbb{E}
     \left[ \ell_{i}\left(\boldsymbol{U}_{\boldsymbol{\theta}\text{, }t},\boldsymbol{U}_{e\text{, }t}\right)+\ell_{c}\left(\boldsymbol{U}_{\boldsymbol{\theta}\text{, }t}\right)\right]$ \ // \ \text{Eq.}\eqref{loss_function}
            \ENDFOR
            \STATE $\boldsymbol{\theta} \gets \boldsymbol{\theta} - \eta \nabla_{\boldsymbol{\theta}} \mathcal{L}$
        \ENDFOR
    \end{algorithmic}
\end{algorithm}

The imitation loss $\ell_i$ is defined as the mean Smooth $L_1$ loss, which provides robustness against outliers while maintaining sensitivity to small errors. Hence, the imitation loss $\ell_i$ is given by 
    \begin{equation}
        \label{loss_function}
\ell_{i}\left(\boldsymbol{U}_{\boldsymbol{\theta}\text{, }t},\boldsymbol{U}_{e\text{, }t}\right) =  SmoothL_1\left(\boldsymbol{U}_{\boldsymbol{\theta}\text{, }t} - \boldsymbol{U}_{e\text{, }t}\right)
    \end{equation}
    where $Smooth L1$ is an element-wise operator defined by
    \begin{equation}
        SmoothL_1(x) = \begin{cases}
        0.5x^2, & \text{if } |x| \leq 1 \\
        |x| - 0.5, & \text{otherwise}
        \end{cases}
    \end{equation}
To encourage the student policy to respect the physical and obstacle-clearance boundaries established in Section \ref{sec:method}, the constraint loss $\ell_c$ is formulated as
\begin{equation}
        \label{loss_function_constr}
\begin{aligned}
\ell_{c}\left(\boldsymbol{U}_{\boldsymbol{\theta}\text{, }t}\right) =&\lambda_{c1}ReLU\left(\underline{\boldsymbol{u}}_{d}-\boldsymbol{A}_{d} \boldsymbol{U}_{\boldsymbol{\theta}\text{, }t}\right)\\
&+\lambda_{c2}ReLU\left(\boldsymbol{A}_{d} \boldsymbol{U}_{\boldsymbol{\theta}\text{, }t}-\overline{\boldsymbol{u}}_{d}\right))\\
&+\lambda_{c3}\left\|\boldsymbol{A}_{c} \boldsymbol{U}_{\boldsymbol{\theta}\text{, }t} -\boldsymbol{c}\right\|_1\\
&+\lambda_{c4}ReLU\left(\boldsymbol{A}_{s}[\boldsymbol{U}_{\boldsymbol{\theta}\text{, }t}^\top, \boldsymbol{0}^\top]^\top - \overline{\boldsymbol{u}}_{s}\right)  
\end{aligned}
\end{equation}
where $\lambda_{i}$ ($i\in\left\{c1\text{, }\ldots\text{, }c4\right\}$) are weights to balance each component. The first two terms penalize dynamic bound violations; the third penalizes $C^2$ continuity across B\'ezier segments; and the final term uses the expert's privileged point cloud and clearance mapping to penalize nominal clearance violations. These terms provides further penalties for the learned policy to satisfy dynamic feasibility, $C^2$ continuity, and collision avoidance.

The PILOT algorithm as shown in Algorithm \ref{pilot_training} follows a two-phase pipeline: a data collection phase and a policy training phase. In the data collection phase, the UAV is randomly initialized in a dense training arena, with trajectories planned by the privileged expert $\pi_e$. In the training phase, the student policy $\pi_{\boldsymbol{\theta}}$ is updated using the Adam optimizer based on the joint loss in \eqref{eq:PIL_Objective}. The hyperparameters for policy training are given in Table \ref{tab:policy_training}.

\begin{table}[tbp]
    \centering
    \caption{Hyperparameters for Policy Training.}
    \setlength{\tabcolsep}{1.0em}{
    \begin{tabular}{lcc}
        \toprule
        \textbf{Parameters} && \textbf{Values} \\
        \midrule
        Imitation loss function   & & $Smooth L_1$ \\
        Type of ResNet          && ResNet18\\
        MLP activation function       & & ReLU \\
        Optimizer                  && Adam \\
        TCN dilation factor       & & $\{1,2,4\}$ \\
        Learning rate $\eta$       && $0.0001$ \\
         MLP hidden units               && 256 \\
        Training epochs        && 200 \\ 
        Batch size                  && 64 \\
        Number of TCN layers    && 3\\
        Number of MLP layers  & & 2\\ 
        \bottomrule
    \end{tabular}
    }
    \label{tab:policy_training}
\end{table}
\section{\textsc{NUMERICAL SIMULATIONS}}\label{sec:sim}
This section presents simulation evaluations of the proposed PILOT framework. We assess planner robustness and generalizability across unseen quadrotor environments, conduct ablation studies on the TCN and trajectory parameterization modules, and demonstrate cross-platform transferability via fixed-wing deployment.

\subsection{Simulation Setup and Configuration}
The PILOT framework is trained on NVIDIA Isaac Sim, leveraging a GPU-accelerated PhysX engine to facilitate large-scale parallel training. Simulations are conducted on a workstation with an Intel Xeon Gold 5218R CPU and an NVIDIA RTX A6000 GPU. The simulated depth stereo camera is modeled after the Intel RealSense D435i, using identical intrinsic and extrinsic parameters to generate the visual data.

\begin{table}[tbp]
    \centering
    \caption{Constraints and hyperparameters for the expert policy}
    \setlength{\tabcolsep}{0.8em}{
    \begin{tabular}{ccc}
        \toprule
        \textbf{Parameters} & \textbf{Quadrotors} & \textbf{Fixed-wing Aircraft} \\
        \midrule
        $\lambda_{g}$ & 8 & 8\\
        $\lambda_{u}$ & 0.4 & 0.4\\
        $\lambda_{s\text{, }1}$ & 70 & 100\\
        $\lambda_{s\text{, }2}$ & 50 & 70\\
        $d_{s}$ & 0.75 & 2.0\\
        $\left\Vert\overline{\boldsymbol{v}}\right\Vert_2$ & 3 & 12 \\
        $\left\Vert\underline{\boldsymbol{v}}\right\Vert_2$ & 0 & 8  \\
        $\left\Vert\overline{\boldsymbol{u}}\right\Vert_2$ & 4 & 16  \\
        $\left\Vert\underline{\boldsymbol{u}}\right\Vert_2$ & 0 & 0  \\
        \bottomrule
    \end{tabular}
    }
    \label{tab:our method's parameters}
\end{table}

The parameters for the MPC expert are detailed in Table~\ref{tab:our method's parameters}, with fixed-wing physical constraints from \cite{Adhikari2020JGCD}. Although the MPC expert uses privileged information, collision-avoidance parameters are tailored to each platform, for example, assigning a larger collision penalty to fixed-wing aircraft due to its reduced agility. The reported parameters follow two selection criteria. Kinematic bounds are dictated by physical platform capabilities, while safety distances and obstacle penalties account for platform scale, maneuverability, perception uncertainty, and turning demands. Objective weights and training hyperparameters are chosen empirically for MPC stability and convergence rather than guaranteed theoretical optimality; temporal history length is backed by the H5/H10/H15 ablation study. To test PILOT's generalization capability, six unseen scenarios with diverse obstacle densities, geometries, and layouts are constructed (Fig.~\ref{fig:mc_trajectory}):
\begin{itemize}
    \item \textbf{Sparse, Moderate, and Dense Arenas}: Incorporate cylindrical obstacles across increasing density ranges ($1/9$–$1/5~\mathrm{obs/m^2}$ for quadrotors; $1/250$–$1/50~\mathrm{obs/m^2}$ for fixed-wing aircraft) with platform-scaled radii.
    \item \textbf{Tilted Cylinder Arena}: Extends the Moderate arena by randomly tilting cylinder $z$-axes in $[0^\circ$ $30^\circ]$.
    \item \textbf{Cubic Arena}: Substitutes cylinders for cubic obstacles ($1/7~\mathrm{cube/m^2}$) to test geometric generalization.
    \item \textbf{Narrow Gap Arena}: Uses gates with a $1.3~\mathrm{m}$ opening to assess navigation under strict spatial bounds.
\end{itemize}

\subsection{Evaluation Metrics}

\textbf{1) Success Rate (SR):}  
The percentage of trials where a UAV reaches the goal without collision. 
\begin{equation*}
    \text{SR} = {N_{s}}\left/{N_{t}}\right. \times 100\%
\end{equation*}  
where \( N_{s} \) is the number of successful trials, and \( N_{t} \) is the total number of trials conducted. In our evaluation, we choose $N_{t}=100$ for all testing scenarios.

\textbf{2) Distance to the Nearest Obstacle (D2NO):}  The mean of the minimum Euclidean distances between the trajectory and obstacle surfaces across successful trials, serving as a measure of safety margin.
\begin{equation*}
    \text{D2NO} = \frac{1}{N_{s}} \sum_{k=1}^{N_{s}} \left( \min_{1 \leq i \leq T_k} \min_{\boldsymbol{o} \subset  \mathcal{X}_o} \| \boldsymbol{p}_k(i) - \boldsymbol{o} \|_2 \right)
\end{equation*}  
where \( T_k \) is the total number of discrete trajectory points in the \(k\)-th successful trial, \( \boldsymbol{p}_k(i) \) denotes the UAV position at the time step \(i\) of the \(k\)-th successful trial.

\textbf{3) Trajectory Length (TL):}  The average total distance traveled during successful missions, indicating path efficiency and the avoidance of unnecessary maneuvers.
\begin{equation*}
    \text{TL} = \frac{1}{N_{s}} \sum_{k=1}^{N_{s}} \sum_{t=1}^{T_k - 1} \| \boldsymbol{p}_k(t+1) - \boldsymbol{p}_k(t) \|_2
\end{equation*}  
A shorter trajectory length implies that the planner is more efficient in generating direct, collision-free trajectories with fewer redundant maneuvers.

\textbf{4) Computation Time (CT):}   The mean processing time per planning step, evaluating the computational cost.
\begin{equation*}
    \text{CT} = \frac{1}{N_{s}} \sum_{k=1}^{N_{s}} \sum_{t=1}^{T_k - 1} t_{k,p}(t)
\end{equation*}  
where $t_{k,p}(i)$ is the computation time at time step $i$ of the $k$-th successful trial. A lower CT value reflects reduced computational latency, enabling faster onboard response times and supporting real-world deployment viability.

\textbf{5) Trajectory Heading Smoothness (THS):}  This metric quantifies the consistency of the aircraft's orientation by evaluating the mean ($\mu_{\text{THS}}$) and variance ($\sigma^2_{\text{THS}}$) of the velocity direction ratios.
\begin{equation*}
    \mu_{\text{THS}} = \frac{1}{N_{s}} \sum_{k=1}^{N_{s}} \left( \frac{1}{T_k-1} \sum_{t=1}^{T_k} \frac{v_{t,y}}{v_{t,x}} \right)
\end{equation*}
\begin{equation*}
    \sigma^2_{\text{THS}} = \frac{1}{N_{s}} \sum_{k=1}^{N_{s}} \left( \frac{1}{T_k-1} \sum_{t=1}^{T_k} \left(\frac{v_{t,y}}{v_{t,x}} - \mu_{\text{THS}}\right)^2 \right)
\end{equation*}
where $\mu_{\text{THS}}$ and $\sigma^2_{\text{THS}}$ denote the mean and variance of the change in direction of the velocity, respectively, and \(v_{t,x}, v_{t,y}\) are the velocity components at the time step \(t\). Lower values correspond to smoother directional changes, leading to more feasible trajectories.

\textbf{6) Trajectory Control Smoothness (TCS):}  Control effort and actuator stress are evaluated by the mean ($\mu_{\text{TCS}}$) and variance ($\sigma^2_{\text{TCS}}$) of the changes in inputs $\boldsymbol{u}$.
\begin{equation*}
    \mu_{\text{TCS}} = \frac{1}{N_{s}} \sum_{k=1}^{N_{s}} \left(\frac{1}{T_k-1} \sum_{i=1}^{T_k-1} \|\boldsymbol{u}_{i+1} - \boldsymbol{u}_i\|_2\right)
\end{equation*}
\begin{equation*}
    \sigma^2_{\text{TCS}} = \frac{1}{N_{s}} \sum_{k=1}^{N_{s}} \left(\frac{1}{T_k-1} \sum_{i=1}^{T_k-1} \left(\|\boldsymbol{u}_{i+1} - \boldsymbol{u}_i\|_2 - \mu_{\text{TC}}\right)^2\right)
\end{equation*}
where $\mu_{\text{TCS}}$ and $\sigma^2_{\text{TCS}}$ denote the mean and variance of acceleration changes, respectively, and $\Delta t$ is the sampling interval. Smaller values indicate smoother variations in controls, suggesting reduced actuator stress and enhanced input efficiency. The THS and TCS metrics together capture both trajectory-level and control-level smoothness.

\subsection{Quadrotor Performance and Ablation Studies}
We evaluate the learned E2E policy on a simulated quadrotor platform whose dynamics and sensor configurations mirror physical system constraints. Low-level tracking relies on a geometric controller \cite{lee2010geometric}. The framework is benchmarked against four primary baselines.
\begin{itemize}
    \item \emph{PILOT-S}: A variant of the proposed framework without TCN, used to demonstrate the necessity of temporal encoding for scene understanding.
    \item \emph{MPC Expert}: A privileged MPC law serving as the performance upper bound.
    \item \emph{EGO-Planner}: An ESDF-free gradient-based local planner based on online B-spline optimization \cite{zhou_ego_2021}.
    \item \emph{NavRL}: A RL-based navigation policy \cite{xu_navrl_2025}  learned via trial-and-error interaction with environments.
\end{itemize}

\subsubsection{Performance Evaluation on Quadrotors}
\begin{figure}[tbph]
    \centering
    \includegraphics[width=0.825\linewidth]{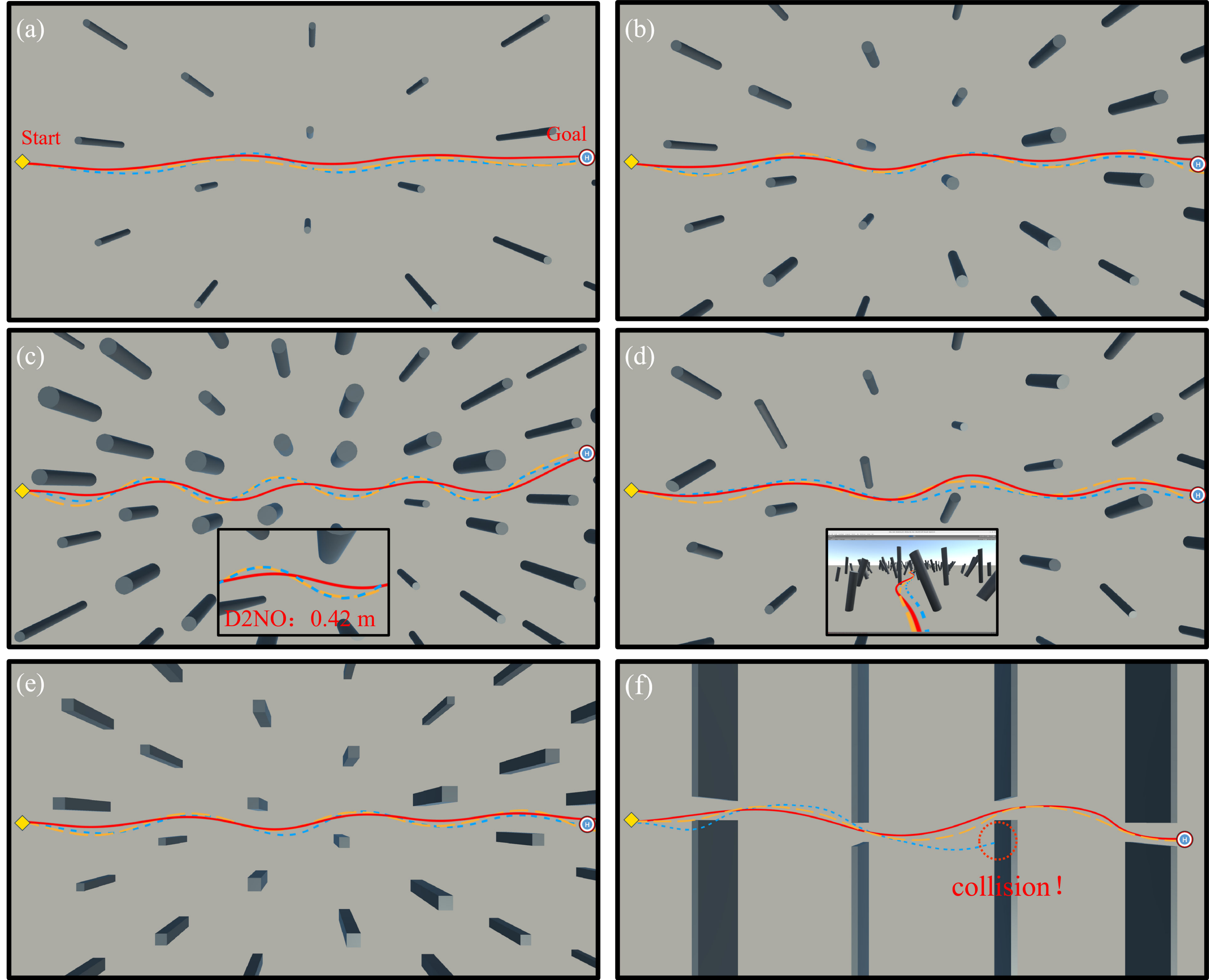}
    \caption{Quadrotor trajectories across evaluation arenas: PILOT (red solid), PILOT-S (blue dotted), and MPC expert (orange dash-dotted).}
    \label{fig:mc_trajectory}
\end{figure}

The evaluation is conducted using Monte Carlo simulations ($100$ trials per arena) across six representative scenarios. The planner performance is assessed using four metrics--such as SR, D2NO, TL, and CT--where Fig. \ref{fig:mc_trajectory} shows the simulated results in one of the testing cases. The overall performance of the policies is summarized in Table \ref{tab:quadrotor_results}. Note that the expert MPC planner is free from visual perception errors, as it uses privileged location information of surrounding obstacles.

\begin{figure}[tbph]
    \centering
    \includegraphics[width=0.9\linewidth]{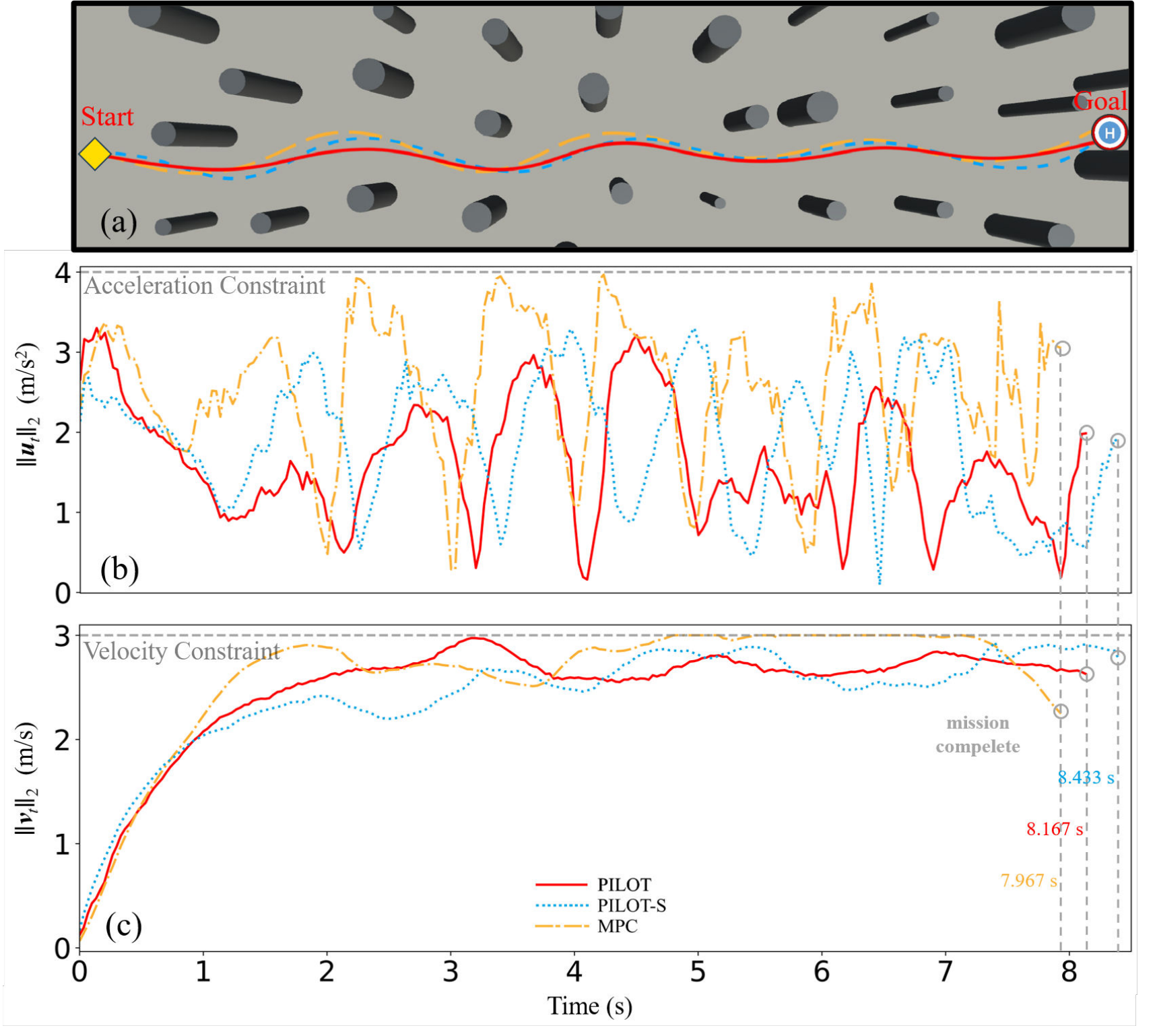}
    \caption{PILOT, PILOT-S, and MPC in the Dense arena.}
    \label{fig:traj_vel_acc_quad}
\end{figure}

\begin{table}[tbph]
\caption{Evaluation of PILOT and benchmarks on quadrotors at different arenas}
\setlength{\tabcolsep}{2pt}
\centering
\label{tab:quadrotor_results}
\label{tab:external_planner_results}
\begin{tabular*}{\linewidth}{@{\extracolsep{\fill}} l l c c c c @{}}
\toprule
Evaluation arenas  & Algorithms & SR (\%) & D2NO (m) & TL (m) & CT (ms)\\
\midrule
 & MPC & 100 & 0.646 & 20.211 & 53.779 \\
 & PILOT & 100 & 0.458 & 20.546 & 8.593 \\
Sparse & PILOT-S & 100 & 0.397 & 20.597 & 4.654 \\
 & EGO-Planner & 100 & 0.449 & 20.531 & 5.756 \\
 & NavRL & 22 & 0.355 & 21.133 & 2.393 \\[0.8ex]

 & MPC & 100 & 0.602 & 20.282 & 54.379 \\
 & PILOT & 100 & 0.424 & 20.559 & 8.816 \\
Moderate & PILOT-S & 93 & 0.418 & 20.684 & 4.341 \\
 & EGO-Planner & 79 & 0.290 & 20.555 & 5.406 \\
 & NavRL & 95 & 0.298 & 21.585 & 2.191 \\[0.8ex]

 & MPC & 100 & 0.519 & 20.860 & 55.200 \\
 & PILOT & 96 & 0.393 & 20.934 & 8.752 \\
Dense & PILOT-S & 82 & 0.350 & 21.127 & 4.596 \\
 & EGO-Planner & 78 & 0.359 & 21.555 & 6.368 \\
 & NavRL & 63 & 0.190 & 21.060 & 2.192 \\[0.8ex]

 & MPC & 100 & 0.603 & 20.279 & 55.671 \\
 & PILOT & 91 & 0.417 & 20.577 & 8.696 \\
Tilted Cylinder & PILOT-S & 83 & 0.409 & 20.646 & 4.691 \\
 & EGO-Planner & 74 & 0.284 & 20.558 & 5.391 \\
 & NavRL & 95 & 0.287 & 21.570 & 2.168 \\[0.8ex]

 & MPC & 100 & 0.603 & 20.280 & 55.827 \\
 & PILOT & 97 & 0.437 & 20.536 & 8.971 \\
Cube & PILOT-S & 93 & 0.422 & 20.624 & 4.927 \\
 & EGO-Planner & 88 & 0.283 & 20.574 & 5.566 \\
 & NavRL & 97 & 0.313 & 21.120 & 2.107 \\[0.8ex]

 & MPC & 100 & 0.679 & 20.635 & 53.046 \\
 & PILOT & 83 & 0.406 & 20.928 & 8.759 \\
Narrow Gap & PILOT-S & 0 & - & - & - \\
 & EGO-Planner & 100 & 0.331 & 20.875 & 6.036 \\
 & NavRL & 81 & 0.183 & 20.856 & 2.168 \\
\bottomrule
\end{tabular*}
\end{table}
As shown in Table \ref{tab:quadrotor_results}, PILOT-S exhibits a lower SR than PILOT in complex scenarios—specifically the Dense, Tilted Cylinder, Cubic, and Narrow Gap arenas. This performance gap stems from the limited FOV in the single-step images used by PILOT-S, whereas PILOT’s TCN-based fusion module implicitly reconstructs privileged obstacle information from temporal sequences to enhance environmental awareness. Furthermore, PILOT achieves consistently higher D2NO values and shorter TL than PILOT-S, yielding efficiency comparable to the MPC expert with significantly reduced computational overhead. Both PILOT and PILOT-S generate velocity and acceleration profiles that respect the quadrotor’s physical constraints, as illustrated in Fig. \ref{fig:traj_vel_acc_quad}. Notably, the MPC expert generates higher accelerations during planning due to a higher navigation cost weight relative to the control input penalty specified in Table \ref{tab:our method's parameters}.

PILOT is also benchmarked against two state-of-the-art methods—EGO-Planner \cite{zhou_ego_2021} and NavRL \cite{xu_navrl_2025}—under identical conditions and metrics. As shown in Table~\ref{tab:external_planner_results}, both PILOT and EGO-Planner attain $100\%$ SR in the Sparse arena. PILOT achieves higher SR in the Moderate, Dense, Tilted Cylinder, and Cube environments, whereas EGO-Planner yields higher SR in the Narrow Gap arena. Among successful trials, EGO-Planner exhibits shorter trajectory lengths and lower computation times, while PILOT maintains larger D2NO values across all arenas. All these indicate that EGO-Planner prioritizes path and computational efficiency, whereas PILOT provides improved navigation success rate and larger empirical obstacle clearances across diverse geometries and densities.

Compared to NavRL, PILOT achieves higher SR across the Sparse, Moderate, Dense, and Narrow Gap arenas, equivalent SR in the Cube arena, and performs slightly lower in Tilted Cylinder. While NavRL consistently yields lower CT values, its reliability varies across environments. This variance stems from NavRL's reliance on 3D ray casting over an occupancy voxel grid  ($10^\circ$ angular resolution in the design). In Sparse arenas featuring thin, isolated obstacles with small angular footprints, obstacles can fall between adjacent rays undetected. In contrast, denser or larger obstacle surfaces are regularly intercepted by rays, yielding a more complete spatial representation. PILOT maintains larger D2NO clearance values and more consistent performance on all arenas. Overall, NavRL prioritizes compact ray-based perception and runtime efficiency, whereas PILOT delivers superior cross-arena reliability and larger empirical safety margins.

To evaluate the impact of historical sequence length ($k$), the baseline PILOT ($k=10$) was compared against PILOT-H5 ($k=5$) and PILOT-H15 ($k=15$) in the Dense arena. As shown in Table~\ref{tab:ablation_sequence_length}, observation history horizon significantly affects performance. PILOT-H5 exhibits a drop in SR to $86\%$ and a reduced minimum obstacle clearance of $0.307~\mathrm{m}$, indicating that insufficient temporal context degrades privileged information reconstruction. Although PILOT-H15 achieves an $89\%$ SR, it still underperforms relative to the baseline. This suggests that increasing sequence length does not monotonically improve performance, as the added capacity complicates policy optimization. Consequently, an intermediate sequence length ($k=10$) provides an effective trade-off between capturing temporal dynamics and maintaining training stability.

\begin{table}[htbp]
    \centering
    \caption{PILOT algorithm performance with different input sequence lengths}
    \setlength{\tabcolsep}{4pt}{
    \begin{tabular*}{\linewidth}{@{\extracolsep{\fill}} l l c c c c c @{}}
        \toprule
        Scenario & Algorithm & SR (\%) & D2NO (m)& TL (m) & CT (ms) \\
        \midrule
         & PILOT-H5   & 100 & 0.383 & 20.553 & 8.362\\
        Sparse & PILOT & 100 & 0.458 & 20.546 & 8.593\\
         & PILOT-H15   & 100 & 0.456 & 20.608 & 8.738 \\[0.8ex]
         
         & PILOT-H5   & 96 & 0.380 & 20.740 & 8.280\\
        Moderate & PILOT & 100 & 0.424 & 20.559 & 8.816\\
         & PILOT-H15   & 97 & 0.409 & 20.618 & 8.503 \\[0.8ex]
         
         & PILOT-H5   & 86 & 0.307 & 21.114 & 8.614\\
        Dense & PILOT & 96 & 0.393 & 20.934 & 8.752\\
         & PILOT-H15   & 89 & 0.329 & 21.127 & 8.782 \\[0.8ex]
        \bottomrule
    \end{tabular*}
    }
    \label{tab:ablation_sequence_length}
\end{table}

\subsubsection{Ablation studies}

\begin{figure}[tbph]
    \centering
    \includegraphics[width=0.95\linewidth]{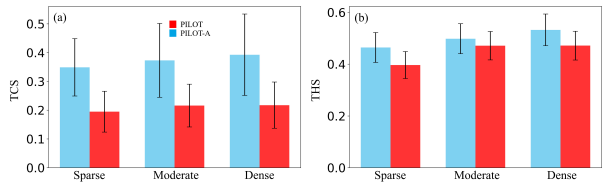}
    \caption{TCS and THS performance of PILOT and PILOT-A based on Monte Carlo simulation ($100$ times for each arena).}
    \label{fig:statistical_data}
\end{figure}

Ablation studies are used to evaluate the trajectory parameterization module by comparing PILOT with PILOT-A, a variant that directly emulates the MPC expert's control inputs. Both models are trained for equal durations and evaluated across Sparse, Moderate, and Dense arenas. As detailed in Table~\ref{tab:ablation_trajectory_parameterization}, PILOT achieves higher success rates and shorter average trajectory lengths than PILOT-A at comparable computational overhead. Analysis of the TCS metric in Fig.~\ref{fig:statistical_data} reveals that PILOT yields smaller mean control input variations ($\mu_{\Delta u}$) and lower variance across all scenarios, reflecting more stable control behavior. This stability is further supported by the THS metric, where PILOT exhibits smoother heading transitions with fewer oscillations than PILOT-A. Collectively, these empirical results indicate that trajectory parameterization improves task success and promotes more executable and stable trajectories in the evaluated cluttered environments.

\begin{table}[tbph]
    \centering
    \caption{Comparison of PILOT and PILOT-A on quadrotors}
    \setlength{\tabcolsep}{4pt}{
    \begin{tabular*}{\linewidth}{@{\extracolsep{\fill}} l l c c c c c c c c c@{}}
        \toprule
        Scenario & Algorithm & SR (\%) & D2NO (m) & TL (m) & CT (ms)\\
        \midrule
        \multirow{2}{*}{Sparse} & PILOT   & 100 & 0.458 & 20.456 & 8.593 \\
                               & PILOT-A & 100 & 0.422 & 20.352 & 8.636\\[0.8ex]
                               
        \multirow{2}{*}{Moderate} & PILOT   & 100 & 0.424 & 20.559 & 8.752 \\
                               & PILOT-A & 93 & 0.413 & 20.438 & 8.556\\[0.8ex]
                               
        \multirow{2}{*}{Dense} & PILOT   & 96 & 0.393 & 20.934 & 8.782 \\
                               & PILOT-A & 90 & 0.358 & 21.129 & 8.753 \\[0.8ex]
        \bottomrule
    \end{tabular*}
    }
    \label{tab:ablation_trajectory_parameterization}
\end{table}

\begin{figure}[btph]
    \centering
    \includegraphics[width=0.925\linewidth]{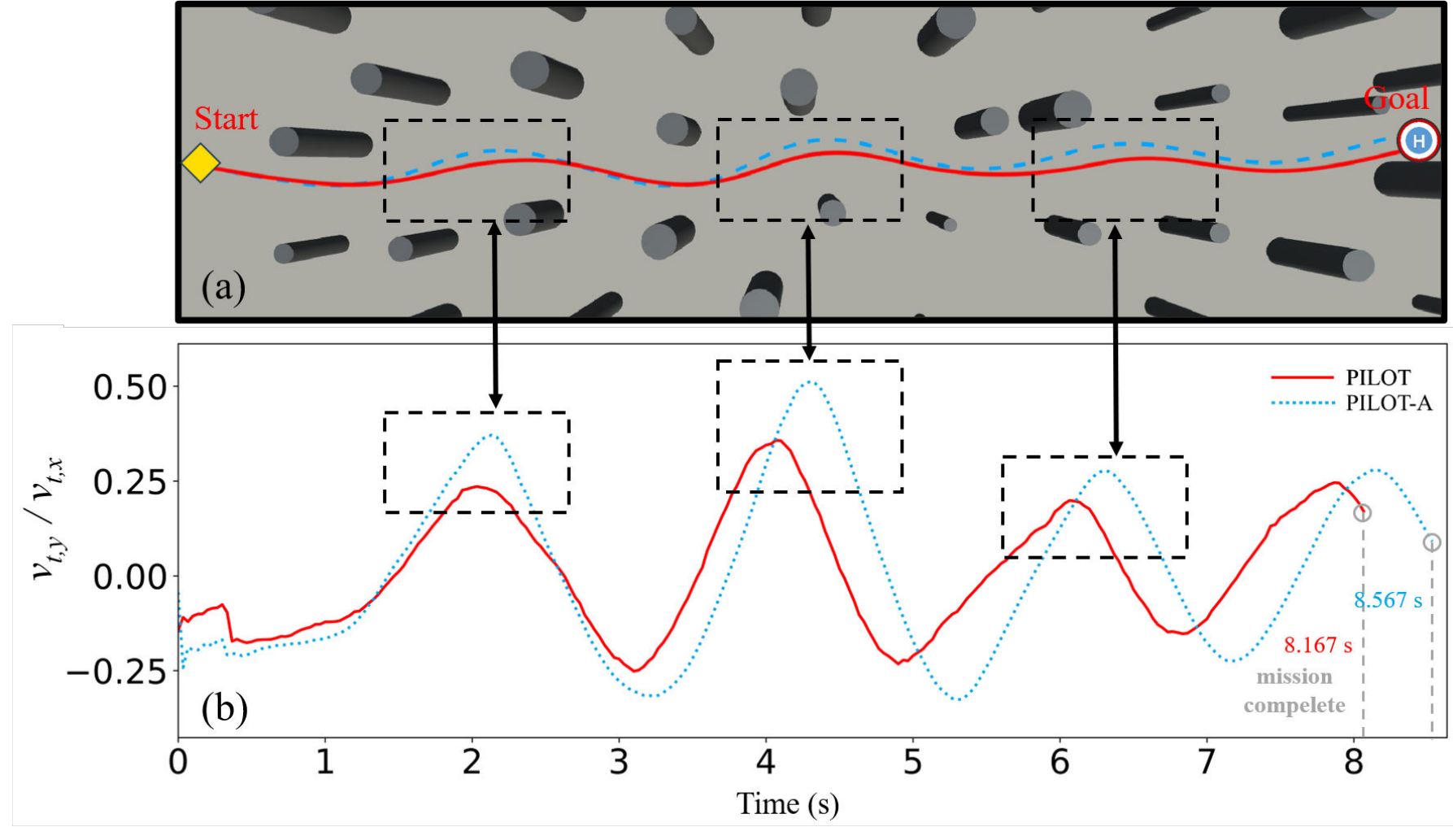}
    \caption{PILOT and PILOT-A in the Dense arena.}
    \label{fig:traj_vx_vy}
\end{figure}

In summary, simulation results and ablation studies show that PILOT learns an efficient visual trajectory planner with high empirical success rates and obstacle clearances in the evaluated cluttered environments. These results characterize performance on the tested distribution and do not constitute a formal runtime safety guarantee. PILOT matches the MPC expert's trajectory quality with reduced computational cost. Additionally, it improves upon PILOT-S across SR, clearance distance, and path efficiency, while outperforming PILOT-A through smoother control profiles enabled by trajectory parameterization.

\subsection{Generalization to Fixed-wing Aircraft}

\begin{table}[btp]
\caption{\label{tab:fixed-wing aircraft simulation results} Comparison of PILOT, PILOT-S, and MPC on fixed-wing aircraft}
\setlength{\tabcolsep}{4pt}
\centering
\begin{tabular*}{\linewidth}{@{\extracolsep{\fill}} l l c c c c @{}}
\toprule
Scenario  & Algorithm & SR (\%) & D2NO (m) & TL (m) & CT (ms) \\
\midrule
 & MPC & 100 & 3.102 & 100.159 & 51.914 \\
Sparse & PILOT & 100 & 2.441 & 100.148 & 8.272 \\
 & PILOT-S & 100 & 2.426 & 100.257 & 4.290 \\[0.8ex]

 & MPC & 100 & 2.041 & 100.238 & 58.734 \\
Moderate & PILOT & 100 & 1.818 & 100.308 & 8.683 \\
 & PILOT-S & 99 & 1.774 & 100.421 & 4.139 \\[0.8ex]

 & MPC & 100 & 1.681 & 100.444 & 51.402 \\
Dense & PILOT & 98 & 1.651 & 100.483 & 8.597 \\
 & PILOT-S & 93 & 1.474 & 100.542 & 4.309 \\[0.8ex]
\bottomrule
\end{tabular*}
\end{table}
To verify the model-agnostic feature of the proposed PILOT framework, we applied it to a fixed-wing aircraft. It is assumed that the fixed-wing aircraft is flying horizontally with its altitude kept constant by a separate controller. Hence, the fixed-wing aircraft dynamics are  
\begin{equation}
\dot{p}_n = V_g \cos{\psi}\text{, }\quad \dot{p}_e= V_g\sin{\psi}
\end{equation}
where $p_n$ and $p_e$ are positions in the north and east directions, respectively; $V_g$ is the ground speed; and $\psi$ is the heading angle.

\begin{figure}[btph]
    \centering
    \includegraphics[width=0.925\linewidth]{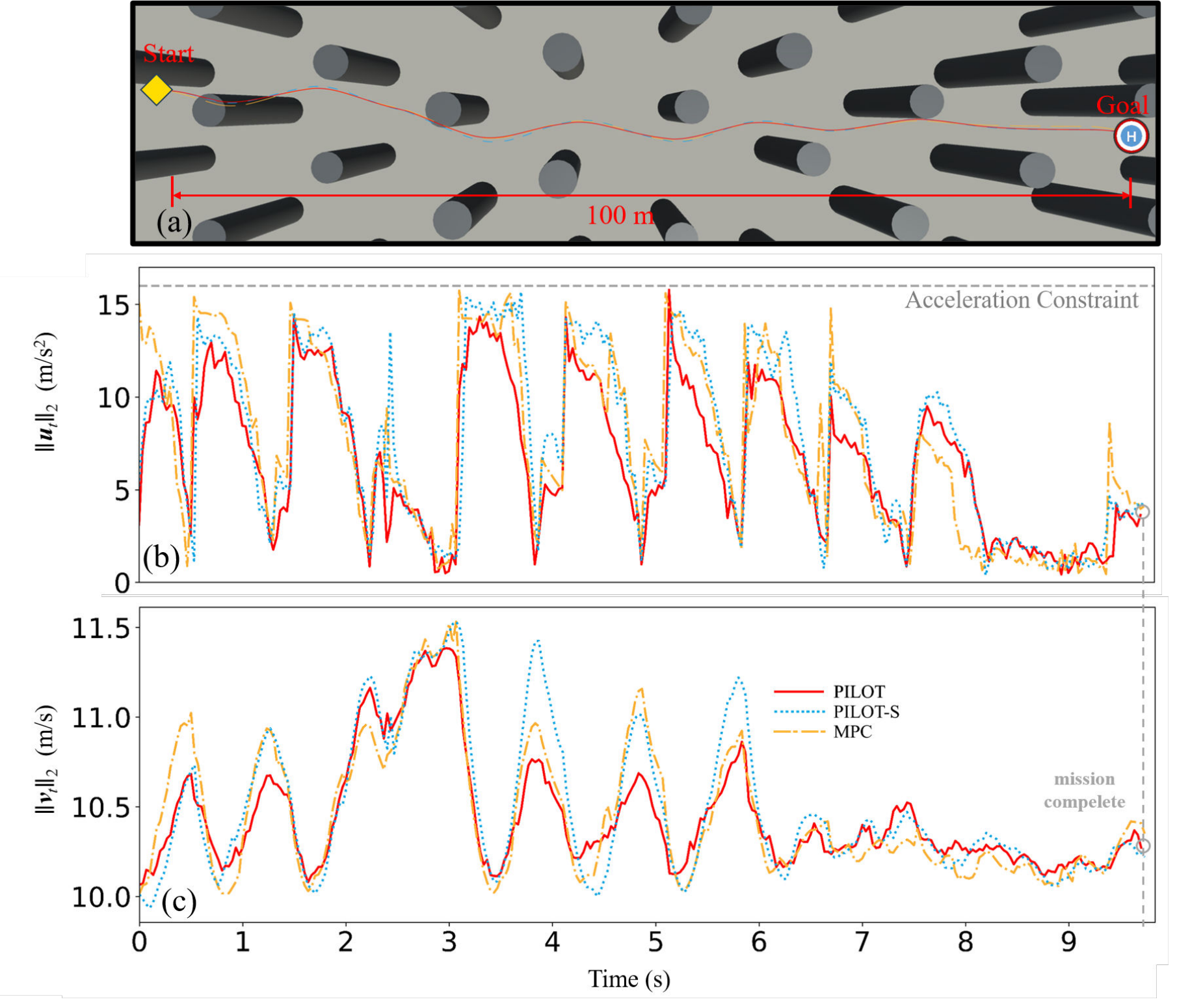}
    \caption{PILOT, PILOT-S, and MPC in the Dense arena (fixed-wing).}
    \label{fig:traj_vel_acc_fix}
\end{figure}

For the evaluation, an L1 guidance law \cite{park2004new} is used to regulate the heading rate by selecting a reference point one second ahead of the current position. As given in Table~\ref{tab:fixed-wing aircraft simulation results}, PILOT achieves performance comparable to the MPC expert across diverse arenas while drastically reducing computational overhead. In Sparse and Moderate environments, PILOT maintains a $100\%$ SR with lower D2NO values than the MPC expert. This indicates that PILOT plans more aggressive trajectories closer to obstacles while retaining a $100\%$ empirical SR in these scenarios; it does not establish safety outside the evaluated conditions. Even in the Dense arena, PILOT maintains a high SR of $98\%$, demonstrating empirical robustness under the tested limited-maneuverability conditions. Compared to the non-temporal baseline (PILOT-S), PILOT attains higher SR and improved D2NO metrics across all environments, further validating the benefit of TCN-based spatiotemporal reasoning. As shown in Fig.~\ref{fig:traj_vel_acc_fix}, both learning frameworks respect fixed-wing constraints, producing path lengths similar to the MPC expert with much less computational latency--making them well-suited for real-time onboard deployment. Overall, these results provide empirical evidence of cross-platform applicability of PILOT.

\section{\textsc{REAL-WORLD EXPERIMENTS}} \label{sec:exp}
\begin{figure}[tbph]
        \centering
    \includegraphics[width=0.75\linewidth]{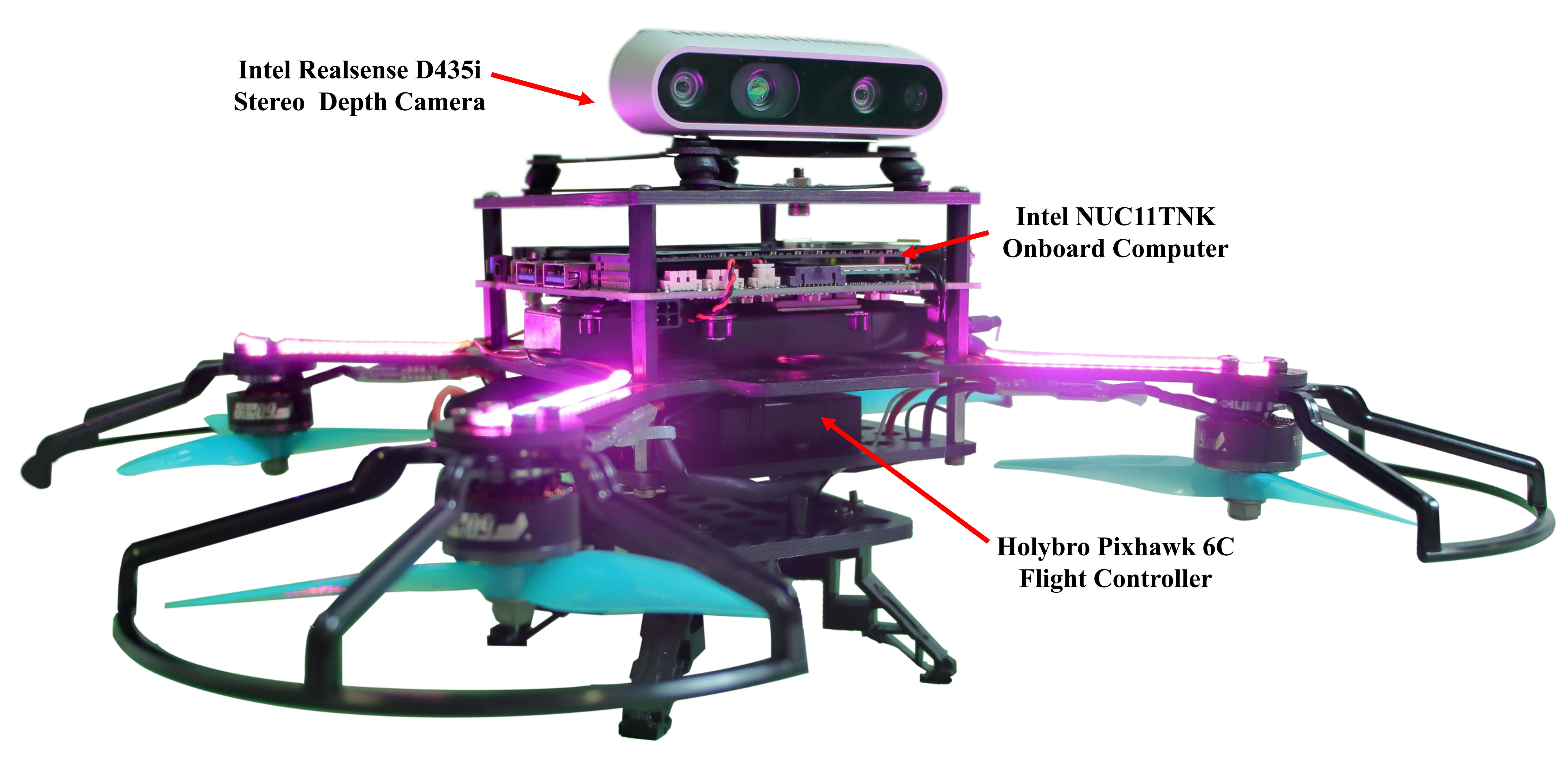}
        \caption{The physical platform used in real-world experiments.}
        \label{fig:Q250}
        \label{fig:quadrotor_platform}
 \end{figure}

To evaluate real-world performance, the simulation-trained PILOT policy was deployed zero-shot onto a custom quadrotor across one indoor and two outdoor unseen environments. As illustrated in Fig.~\ref{fig:quadrotor_platform}, the autonomous platform features an Intel RealSense D435i stereo camera ($\approx 7~\mathrm{m}$ range) for depth sensing and VINS-Fusion for visual-inertial odometry. High-level localization and planning run in real time on an onboard Intel NUC11TNK computer with low-level attitude control provided by PX4.

\begin{figure}[btph]
    \centering
    \includegraphics[width=0.9\linewidth]{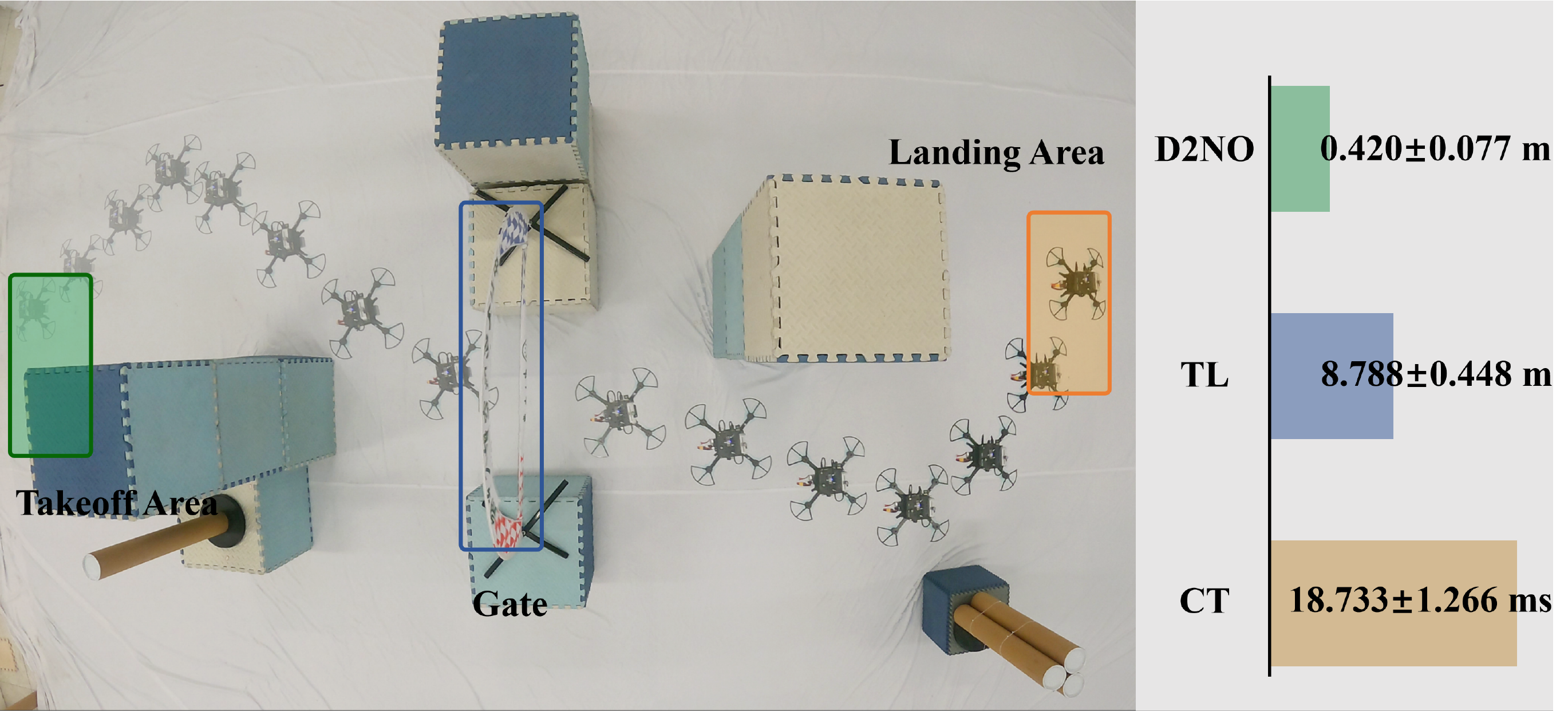}
    \caption{Quadrotor flight trajectory in the indoor experiment.}
    \label{fig:indoor-exp}
\end{figure}

For the indoor evaluation, the quadrotor navigates a constrained $5~\mathrm{m} \times 9~\mathrm{m}$ space containing rectangular prisms and doorways to reach a goal $6.5~\mathrm{m}$ away. The policy operates under \emph{zero-shot transfer}, receiving no real-world fine-tuning. As depicted in Fig.~\ref{fig:indoor-exp}, the learned planner generates a trajectory online that is successfully tracked through the densely arranged obstacles using onboard perception. This experiment demonstrates zero-shot transfer and practical feasibility in the tested indoor environment, rather than a formal guarantee of dynamic feasibility or robustness under arbitrary real-world uncertainties.

\begin{figure}[tbph]
    \centering
    \includegraphics[width=0.9\linewidth]{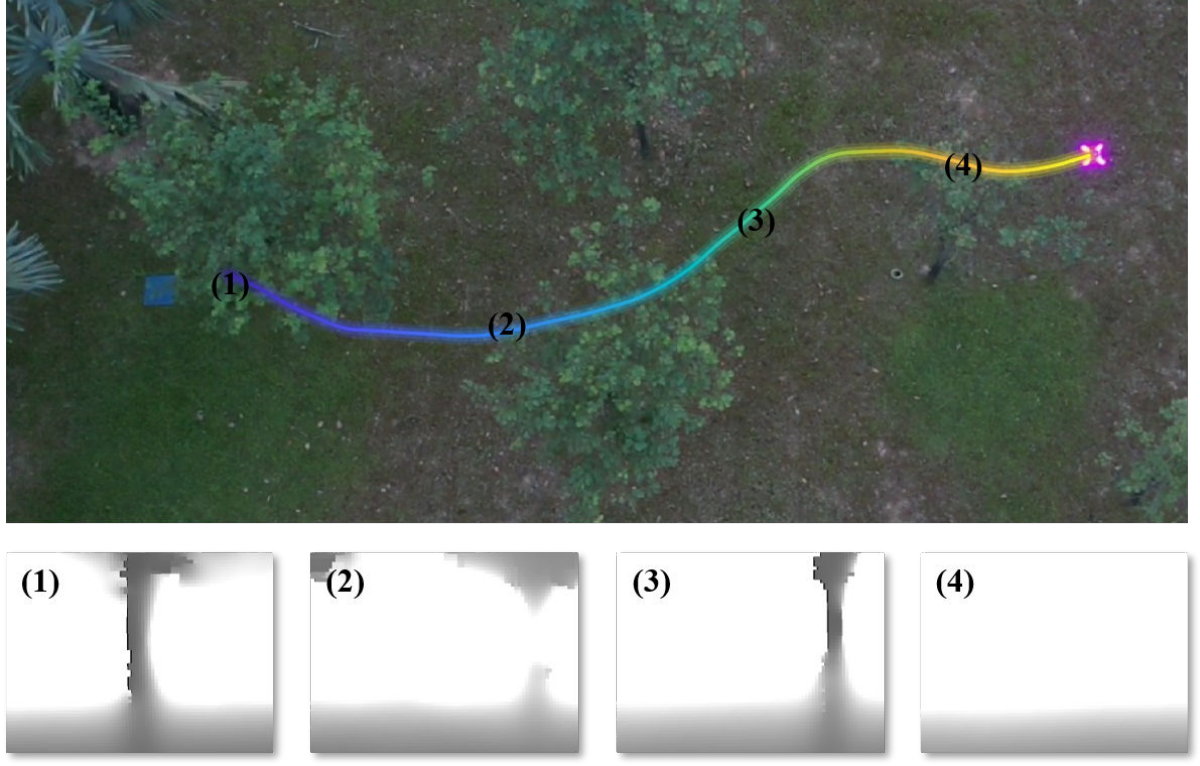}
    \caption{Experiments in a wooded area: top figure shows the flight trajectory, while (1)–(4) showing visual perception at key positions. }
    \label{fig:real trajectory wooded}
\end{figure}

\begin{figure}[tbph]
    \centering
    \includegraphics[width=0.9\linewidth]{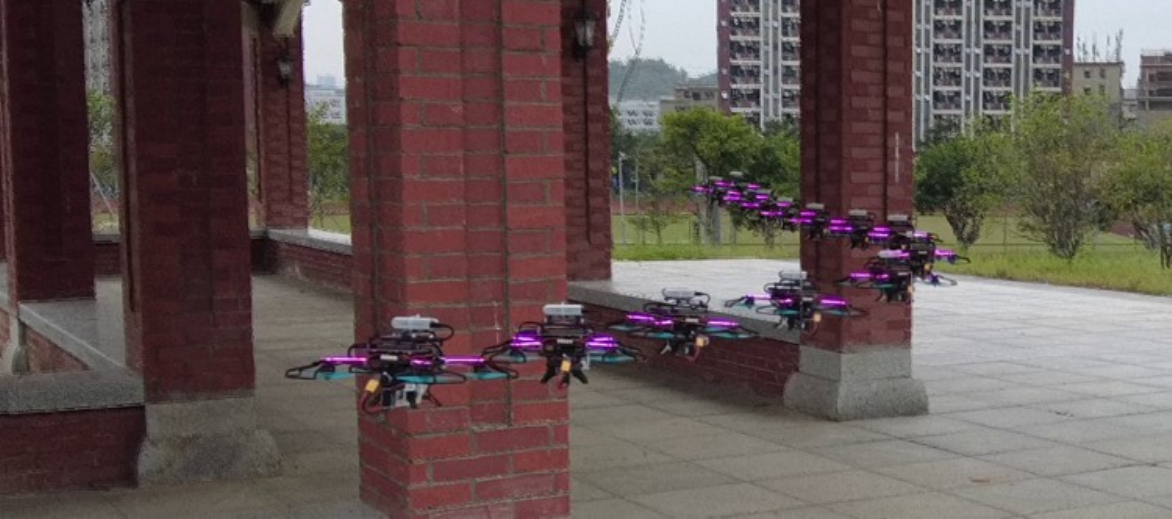}
    \caption{Flight experiments in environments with pillars, demonstrating the generalization of the learned E2E planner.}
    \label{fig:real trajectory pillar}
\end{figure}

In the first outdoor experiment, the quadrotor equipped with the learned E2E planner is deployed in a natural wooded environment, as shown in Fig.~\ref{fig:real trajectory wooded}. The learned planner successfully navigate the quadrotor through wooded area over a distance of approximately $15$~m to the target point. The obstacle-avoidance behavior relies exclusively on onboard visual perception. This result demonstrates practical feasibility in the wooded environment. In the second outdoor experiment, the proposed method is evaluated in an environment with pillar obstacles, as shown in Fig.~\ref{fig:real trajectory pillar}. These demonstrations provide evidence of zero-shot transfer across real-world environments.

\section{\textsc{CONCLUSIONS}}
\label{sec:conclusion}
In this paper, we proposed PILOT (Privileged Imitation Learning for vision-based E2E UAV mOTion planning), a learning algorithm that trains an E2E planner for UAV navigation in cluttered environments using limited visual measurements and constrained onboard computation.
Demonstration data generated by a privileged MPC expert guides offline policy training. To bridge the information gap, a perception fusion module based on a TCN infers task-relevant context from finite spatiotemporal measurements, while a trajectory parameterization module provides a structured output and supports smoothness and constraint-aware regularization. The learned planner is evaluated through simulations and hardware flight tests, demonstrating it matches the MPC expert's trajectory quality while cutting computational latency by over $80\%$. Ablation studies confirm the necessity of both the TCN and trajectory parameterization components. Real-world experiments demonstrate zero-shot transfer and collision-free completion in the reported trials. Because the student uses training-time soft penalties, these results provide empirical evidence in the tested environments rather than a formal safety guarantee for arbitrary unseen conditions. Future work will incorporate dynamic obstacles and external disturbances to enhance generality and robustness.

\bibliography{pilot_refs} %
\bibliographystyle{IEEEtran} 

\end{document}